\documentclass[journal]{IEEEtai}

\usepackage[colorlinks,urlcolor=blue,linkcolor=blue,citecolor=blue]{hyperref}

\usepackage{color,array}

\usepackage{algpseudocode}
\usepackage[ruled, vlined, linesnumbered]{algorithm2e}
\usepackage{hyperref}
\usepackage{cprotect}
\usepackage{graphicx}
\usepackage{textcomp}
\usepackage{xcolor}
\usepackage{tikz}
\usepackage{subcaption}
\usepackage{multirow}
\usepackage{booktabs}
\usepackage{amsmath}
\usepackage{mathtools}
\usepackage{amssymb}
\usepackage[table]{xcolor}
\usepackage{changes}

\newcommand{\qdist}[1]{\ifmmode\langle#1\rangle\else\textlangle#1\textrangle\fi}

\begin{document}

\title{FedSub: Introducing Class-aware Subnetworks Fusion to Enhance Personalized Federated Learning} 

\author{Mattia Giovanni Campana, and Franca Delmastro
\thanks{M. G. Campana is with the Institute for Informatics and Telematics (IIT), National Research Council of Italy (CNR), Italy (e-mail: mattia.campana@iit.cnr.it).}
\thanks{F. Delmastro is with the Institute for Informatics and Telematics (IIT), National Research Council of Italy (CNR), Italy (e-mail: franca.delmastro@iit.cnr.it).}}

\markboth{Journal of IEEE Transactions on Artificial Intelligence, Vol. 00, No. 0, Month 2020}
{First A. Author \MakeLowercase{\textit{et al.}}: Bare Demo of IEEEtai.cls for IEEE Journals of IEEE Transactions on Artificial Intelligence}

\maketitle

\begin{abstract}
Personalized Federated Learning aims at addressing the challenges of non-IID data in collaborative model training. However, existing methods struggle to balance personalization and generalization, often oversimplifying client similarities or relying too heavily on global models.
In this paper, we propose FedSub, a novel approach that introduces class-aware model updates based on data prototypes and model subnetworks fusion to enhance personalization. Prototypes serve as compact representations of client data for each class, clustered on the server to capture label-specific similarities among the clients. Meanwhile, model subnetworks encapsulate the most relevant components to process each class and they are then fused on the server based on the identified clusters to generate fine-grained, class-specific, and highly personalized model updates for each client.
Experimental results in three real-world scenarios with high data heterogeneity in human activity recognition and mobile health applications demonstrate the effectiveness of FedSub with respect to state-of-the-art methods to achieve fast convergence and high classification performance.
\end{abstract}

\begin{IEEEImpStatement}

FedSub introduces a new paradigm for Personalized Federated Learning through the use of class-aware subnetworks.
Traditional approaches strongly rely on global model aggregation or coarse client clustering, which often fail to capture the fine-grained heterogeneity and the complexity of real-world data.
FedSub, instead, identifies the most relevant subnetworks for each class and selectively fuses them across clients, enabling model updates that are both highly personalized and structurally efficient.
This design not only enhances accuracy and convergence under non-IID data distributions, but also offers architectural advantages: clients can update and exchange only class-specific subnetworks rather than entire models, potentially reducing communication costs and enhancing scalability in resource-constrained environments.
By advancing a subnetwork-based perspective, FedSub strengthens the technological foundations of distributed AI, paving the way for more adaptive, efficient, and privacy-preserving learning systems in scenarios where personalization is critical.
\end{IEEEImpStatement}


\section{Introduction}
\label{sec:intro}

Federated Learning (FL) has emerged as a powerful solution for training machine learning models in distributed environments. In this setting, data remains on the clients, and only model updates are shared with a central server to collaboratively learn a single global model~\cite{ZHANG2021106775}.
Despite its potential when training data is scarce on individual devices or privacy is a key requirement, FL faces significant challenges when clients' data are \emph{non-independent and identically distributed} (\emph{non-IID}).
This characteristic is common in real-world applications, where user data reflects differences in behavior, preferences, or physiological traits, and standard FL approaches such as FedAvg~\cite{pmlr-v54-mcmahan17a} typically under-perform, since a global average model cannot capture the diverse and personalized patterns of each user.

To overcome this limitation, Personalized Federated Learning (PFL) has been proposed to address data heterogeneity by tailoring models to individual clients~\cite{9743558}, and it has shown successful applications in various domains, including personalized healthcare~\cite{10.1145/3501296}, smart home automation~\cite{Yang2023}, meteorological sensor networks~\cite{SU2024}, driver-vehicle interaction~\cite{10.1145/3631421}, and multi-device environments~\cite{10.1145/3550289}.

\begin{figure*}[t]
     \centering
     \begin{subfigure}[t]{0.24\textwidth}
         \centering
         \includegraphics[height=3.5cm, width=\linewidth]{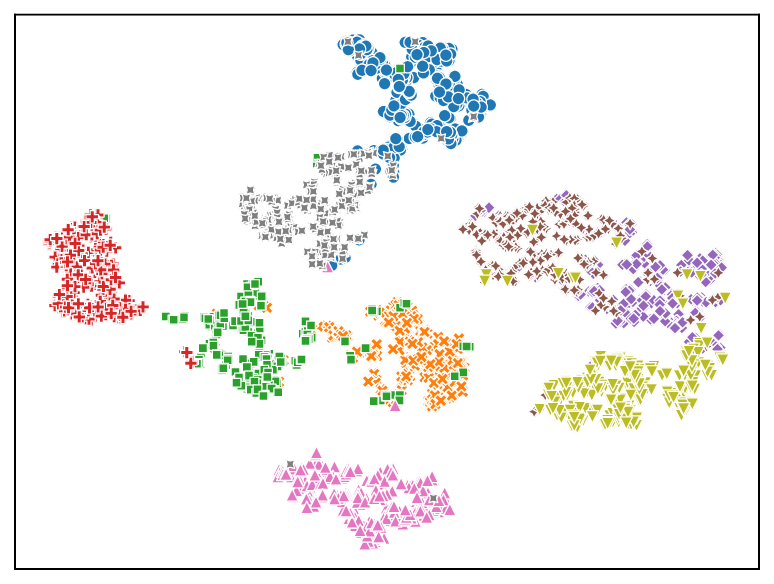}
         \caption{walk}
         \label{fig:walking_acc}
     \end{subfigure}
     \hfill
     \begin{subfigure}[t]{0.24\textwidth}
         \centering
         \includegraphics[height=3.5cm, width=\linewidth]{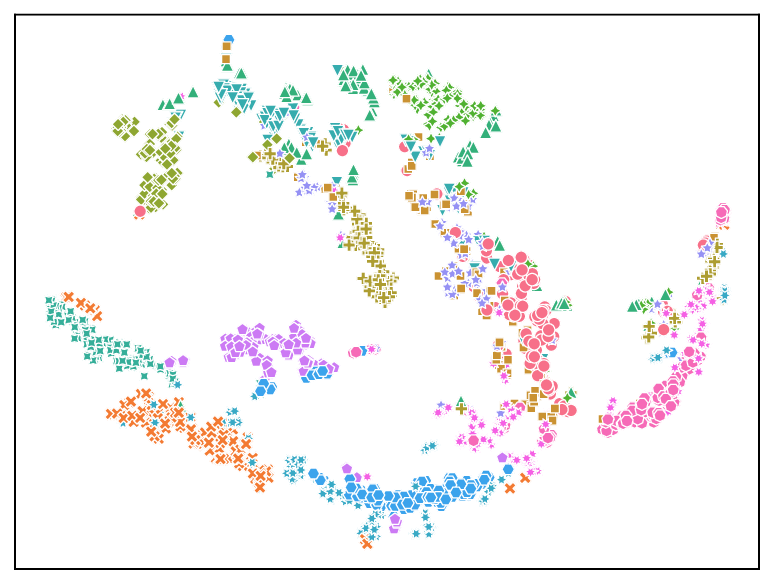}
         \caption{stress}
         \label{fig:stress_ecg}
     \end{subfigure}
     \hfill
     \begin{subfigure}[t]{0.49\textwidth}
         \centering
         \includegraphics[height=3.5cm, width=\linewidth]{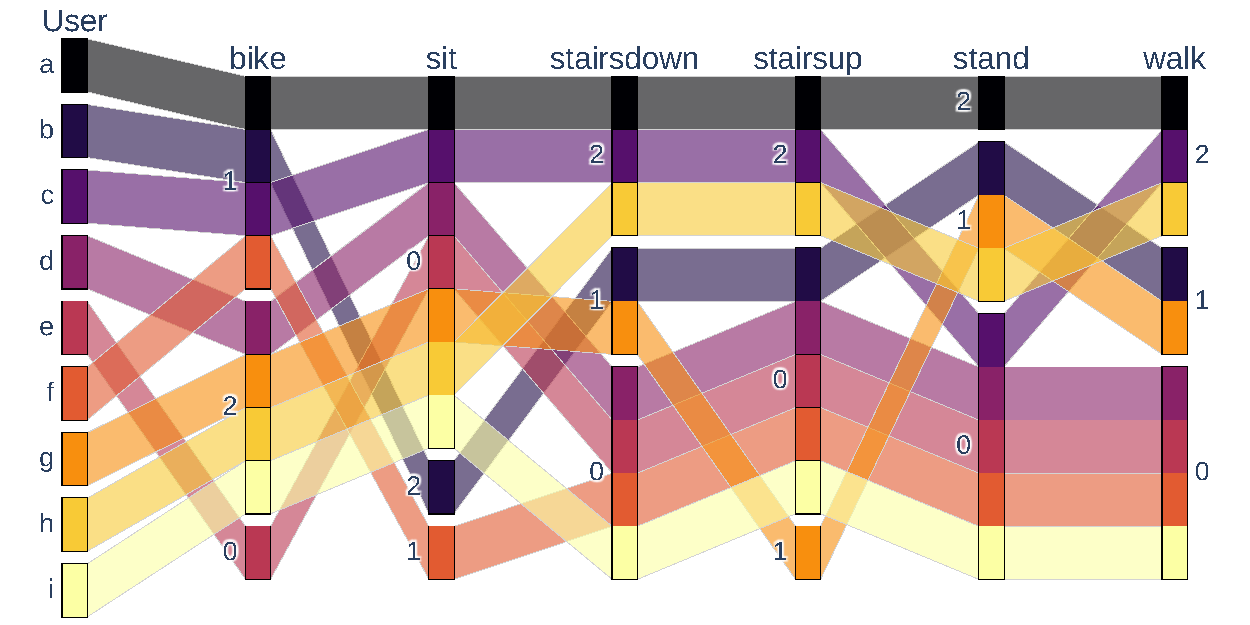}
         \caption{Class-level clusters in HHAR}
         \label{fig:hhar_clusters}
     \end{subfigure}
    \caption{Specificity of users' data patterns. Figures a) and b) display the t-SNE representations of user patterns for the same data classes in HHAR and WESAD, respectively, while Figure c) highlights the variation in user clusters across the different labels available in HHAR.}
    \label{fig:non-iid_examples}
\end{figure*}

A common approach to implement PFL is to adapt the global model through regularized local fine-tuning. However, this strategy struggles to balance personalization and generalization, especially at scale, since the global model may fail to capture client-specific patterns~\cite{10.1145/3554980}.
In contrast, other approaches adopt different clustering techniques to train distinct global models for groups of clients with similar data characteristics~\cite{10.1145/3580795}.
Nevertheless, the assumption that parameter similarity directly corresponds to data similarity is overly simplistic, as clients may share common patterns only for certain classes while differing significantly for others, ultimately resulting in incomplete personalization.

To illustrate the limitations of current PFL approaches and highlight the motivations of our work, we focus on mobile health (mHealth) and Human Activity Recognition (HAR) applications, where non-IID data issues are particularly pronounced.
Indeed, in HAR, different users may perform the same activity in various ways due to differences in age, habits, or physical traits, while in mHealth, physiological signals can vary even within the same user category, depending on individual responses to stimuli or health conditions. These characteristics make such applications ideal testbeds for PFL.

We analyze the clustering tendency at the class level of two real datasets as an example of our assumption: Heterogeneity Human Activity Recognition (HHAR)~\cite{10.1145/2809695.2809718}, which contains accelerometer data of 6 users while performing common activities such as walking and riding a bike, and Wearable Stress and Affect Detection (WESAD)~\cite{10.1145/3242969.3242985}, a multimodal dataset that contains physiological and motion data collected during a clinical protocol on stress detection from 15 subjects.
Specifically, we calculate the Hopkins statistic~\cite{QIU2016216} at the class level, comparing the distance between data points of each class with those of randomly generated points.
A value close to $1$ indicates a strong tendency to clustering, while a result close to $0$ suggests that data points are uniformly distributed.

\begin{table}[b]
\caption{Accuracy performance before and after combining the models of users A and B from HHAR.}
\label{tab:merge_test}
\centering
\begin{tabular}{l|rr|rr}
\toprule
\textbf{Class} & \textbf{A/A} & \textbf{B/B} & \textbf{A+B/A} & \textbf{A+B/B} \\
\midrule
stand      & 0.78    & 0.99    & \textcolor{red}{0.00 ($- 0.78 \downarrow$)}        & \textcolor{red}{0.00 ($- 0.99 \downarrow$)}        \\
sit        & 1.00    & 1.00    & \textcolor{red}{0.00 ($- 1.00 \downarrow$)}        & \textcolor{red}{0.00 ($- 1.00 \downarrow$)}        \\
walk       & 0.75    & 0.87    & \textcolor{red}{0.00 ($- 0.75 \downarrow$)}        & \textcolor{red}{0.81 ($- 0.06 \downarrow$)}        \\
stairsup   & 0.87    & 1.00    & \textcolor[HTML]{28803f}{1.00 ($+ 0.13 \uparrow$)}        & \textcolor{red}{0.00 ($- 1.00 \downarrow$)}        \\
stairsdown & 0.34    & 0.66    & \textcolor{red}{0.30 ($- 0.04 \downarrow$)}        & \textcolor[HTML]{28803f}{0.82 ($+ 0.16 \uparrow$)}        \\
bike       & 0.90    & 0.97    & \textcolor{red}{0.36 ($- 0.54 \downarrow$)}        & \textcolor[HTML]{28803f}{1.00 ($+ 0.03 \downarrow$)}       \\
\bottomrule
\end{tabular}
\end{table}

We obtained values between 0.89 and 0.99 for both datasets, indicating that user data for the same class tend to form distinct clusters.
This is further visualized through t-SNE~\cite{van2008visualizing} in Figures~\ref{fig:walking_acc} and~\ref{fig:stress_ecg}, which show the 2-D representations of data points for the \verb|walk| (HHAR) and \verb|stress| (WESAD) classes.
These results highlight the variability in user data, suggesting that a single global model would be inefficient and could lead to suboptimal performance.

Moreover, clustering entire models can also be problematic.
Figure~\ref{fig:hhar_clusters} shows how users in HHAR are clustered, noting that their cluster membership significantly changes across different activities. For example, users \verb|A| and \verb|B| exhibit similar patterns for \verb|bike| (i.e., they belong to the same cluster, \verb|1|), but differ in other activities, which means that combining their models based solely on their shared cluster could deteriorate the model's ability to recognize other activities.

To test this hypothesis, we trained a basic feedforward neural network on each user's local data, then we merged the models of users \verb|A| and \verb|B|, evaluating the combined model on their respective test sets. Table~\ref{tab:merge_test} shows the classification accuracy for each class, where, for example, \textbf{A+B/A} denotes the performances of the model obtained by averaging the models of \verb|A| and \verb|B|, tested on \verb|A|'s test set.

The results reveal that merging models with different activity patterns is generally disruptive. Indeed, the distinct users' models recognized \verb|stand| and \verb|sit| with accuracies of $0.78$ and $0.99$ but, after merging, the accuracy dropped to $0$ for both.
In some cases, the merging operation brings some advantages, as in \verb|stairsdown| for \verb|B| that passes from $0.66$ to $0.82$.
However, for their shared activity \verb|bike|, performance dropped for \verb|A| (from $0.90$ to $0.36$), while it reached $100\%$ for \verb|B|, reflecting subtle differences in their patterns despite being in the same cluster.
These results demonstrate that simply merging entire models, even for clients in the same cluster, could be inefficient, highlighting the need for more personalized approaches.

\subsection{Contribution}

As we highlighted in the previous example, the conventional strategy of combining full models on the server can lead to suboptimal performance, especially when data is strongly heterogeneous across clients.
Global updates tend to weaken client-specific information, while local fine-tuning alone fails to fully exploit shared patterns, thus resulting in poor personalization and slow or unstable convergence.

To address these limitations, we propose \textbf{FedSub}, a novel algorithm that introduces a radically different paradigm for model fusion.
It leverages two key concepts: \emph{class-aware prototypes} and \emph{subnetwork fusion}. Class-aware prototypes provide compact class-level representations of each client’s data, clustered on the server to reveal similarities across users. Subnetworks, inspired by Explainable AI~\cite{10.1371/journal.pone.0130140} and model pruning~\cite{YEOM2021107899}, capture the most relevant parameters for each class and are then selectively fused according to prototype clusters. This design enables personalized updates that reflect both shared patterns among similar users and client-specific characteristics.

In this work, we evaluate the effectiveness of our proposal from several perspectives:

\begin{itemize}

\item \emph{Adaptive personalization}: we design and compare two different mechanisms to extract class-level subnetworks on the client, and three policies to fuse them on the server.

\item \emph{Performance enhancement with respect to the state-of-the-art}: through extensive experiments on three real-world datasets and two data generation scenarios, we show that FedSub consistently improves accuracy and convergence over state-of-the-art PFL methods in heterogeneous mobile and wearable settings.

\item \emph{Scalability analysis}: we provide an in-depth analysis of FedSub complexity, showing that it is scalable with respect to the number of clients, and computationally efficient on the client side.

\item \emph{Sensitivity analysis}: we assess the impact of subnetworks extraction and fusion policy on both personalization performance and communication overhead.

\end{itemize}

Moreover, to support and foster further research in PFL, we publicly release the FedSub implementation, together with the simulation code, preprocessing scripts, and datasets at \cprotect{\href{https://anonymous.4open.science/r/FedSub}}{\verb|https://anonymous.4open.science/r/FedSub|}.

The remainder of the manuscript is organized as follows. Section~\ref{sec:related} reviews key PFL approaches, their limitations, and how FedSub addresses them. Section~\ref{sec:methodology} details the architecture design and functionalities.
Section~\ref{sec:experiments} presents the experimental evaluation, covering datasets preparation and baseline comparisons. In Section~\ref{sec:hyperparams}, we analyze the impact of FedSub's variations, focusing on how subnetwork definitions and fusion strategies can impact both communication overhead and classification performance. Finally, in Section~\ref{sec:conclusions}, we draw our conclusions and present some directions for future work.

\section{Related Work}
\label{sec:related}

To address the challenges posed by non-IID data and enable personalization in FL, several solutions have been proposed in the last few years.
According to a recent survey~\cite{9743558}, they can be grouped into two classes: (i) data-centric and global model personalization, and (ii) client-specific model learning.

\subsection{Data-centric and Global model personalization}

A first line of research focuses on data-centric methods, such as \emph{data augmentation}.
FAug~\cite{9220773}, for example, uses Generative Adversarial Networks (GANs) to generate synthetic samples for underrepresented classes, with the aim of rebalancing the data distributions before fine-tuning.
Similarly, FedHome~\cite{9296274} uses SMOTE~\cite{chawla2002smote} to oversample minority classes in home health monitoring applications. Both approaches rely on strong assumptions about the availability or quality of synthetic data, which may not hold in privacy-sensitive or resource-constrained environments.

Another line of work is represented by \emph{regularization-based} methods that aim to stabilize training and improve personalization by constraining local models toward the global one.
The most relevant ones are the following.
FedProx~\cite{MLSYS2020_1f5fe839} penalizes local model divergence from the global model. Ditto~\cite{pmlr-v139-li21h} jointly learns both a global and a specific model for each client by using a proximal penalty as regularization term to balance the use of global knowledge with the need to adapt to individual client data. SCAFFOLD~\cite{pmlr-v119-karimireddy20a} introduces control variates to mitigate client drift, and MOON~\cite{Li_2021_CVPR} leverages contrastive learning to align local and global representations.
pFedMe~\cite{NEURIPS2020_f4f1f13c} employs the Moreau envelope to balance local personalization with global alignment.
While effective, these methods restrict personalization to remain close to the global optimum, limiting flexibility under highly non-IID data.

Meta-learning strategies such as Per-FedAvg~\cite{NEURIPS2020_24389bfe} leverage Model-Agnostic Meta-Learning (MAML)~\cite{pmlr-v70-finn17a} to learn a global model initialization that clients can adapt with a few local gradient steps.
While this improves adaptability, it requires expensive second-order gradient computations~\cite{9743558}, which can be prohibitive for resource-constrained devices.

Recently, \emph{prototype-based} methods have emerged as a simple, yet effective alternative in the area of global model personalization. FedProto~\cite{Tan_Long_LIU_Zhou_Lu_Jiang_Zhang_2022} aggregates class-wise prototypes instead of model weights to provide a global representation of the data, which are then used to regularize local training aimed at improving alignment between client models, but without altering the way global aggregation is performed.
Similarly, ProtoHAR~\cite{10122911} exchanges prototypes of class representations and aligns local models through a shared representation network.
The main drawback of current prototypes-based approaches lies in the reliance on a single global representation, which tends to mask inherent differences in local data distributions and prevents class-specific personalization.

\subsection{Client-specific model learning}

This second strategy aims to redesign the FL framework to directly produce individualized models.
One popular approach is \emph{parameter decoupling}, in which models are split into shared and private parts (e.g., features extraction and classifier): while shared components are aggregated globally on the server, private parameters are trained locally to enhance personalization~\cite{arivazhagan2019federated}.
Although conceptually simple, these methods require careful tuning of the split and do not adapt to class-specific needs in each client’s distribution.

Another relevant research direction involves \emph{clustering} clients with similar data distributions. For example, CFL~\cite{9174890} clusters clients based on the cosine similarity of their gradients, while Briggs et al.~\cite{9207469} proposed to use hierarchical clustering based on models divergence.
In healthcare applications, FedCLAR~\cite{9762352} and CBFL~\cite{HUANG2019103291} group users based on model parameters or latent features, and train a dedicated model for each cluster.
Their main weakness lies in the assumption that similarities at the model level (gradients or weights) directly reflect similarities in client data. In practice, clients often overlap only partially, making full models aggregation suboptimal and hindering finer-grained personalization.

\subsection{FedSub positioning}

FedSub mainly falls within the second category, but it introduces a new paradigm that goes beyond the previous solutions.
Specifically, unlike parameter-decoupling techniques, FedSub extracts class-aware subnetworks that encapsulate the most relevant parameters for each class. These subnetworks are semantically meaningful rather than arbitrarily partitioned, enabling more targeted knowledge sharing while avoiding the transmission of irrelevant parameters.
 
At the same time, FedSub differs from prototype-based methods such as FedProto and ProtoHAR by leveraging prototypes to capture class-level similarities across clients, which also serve as structural guidance for selective subnetwork fusion.
In this way, prototypes enable fine-grained, class- and client-specific updates that integrate only the most relevant knowledge, instead of merely enforcing consistency during local optimization.
By combining similarity guidance with subnetwork-based aggregation, FedSub overcomes the limitations of both approaches and achieves stronger personalization with more stable convergence in heterogeneous scenarios.

\newcommand*\circled[1]{\tikz[baseline=(char.base)]{
            \node[shape=circle,draw,inner sep=1pt] (char) {\footnotesize #1};}}

\begin{figure}[t]
     \centering
    \begin{tikzpicture}
        \node [inner sep=0pt,above right]{\includegraphics[width=\columnwidth]{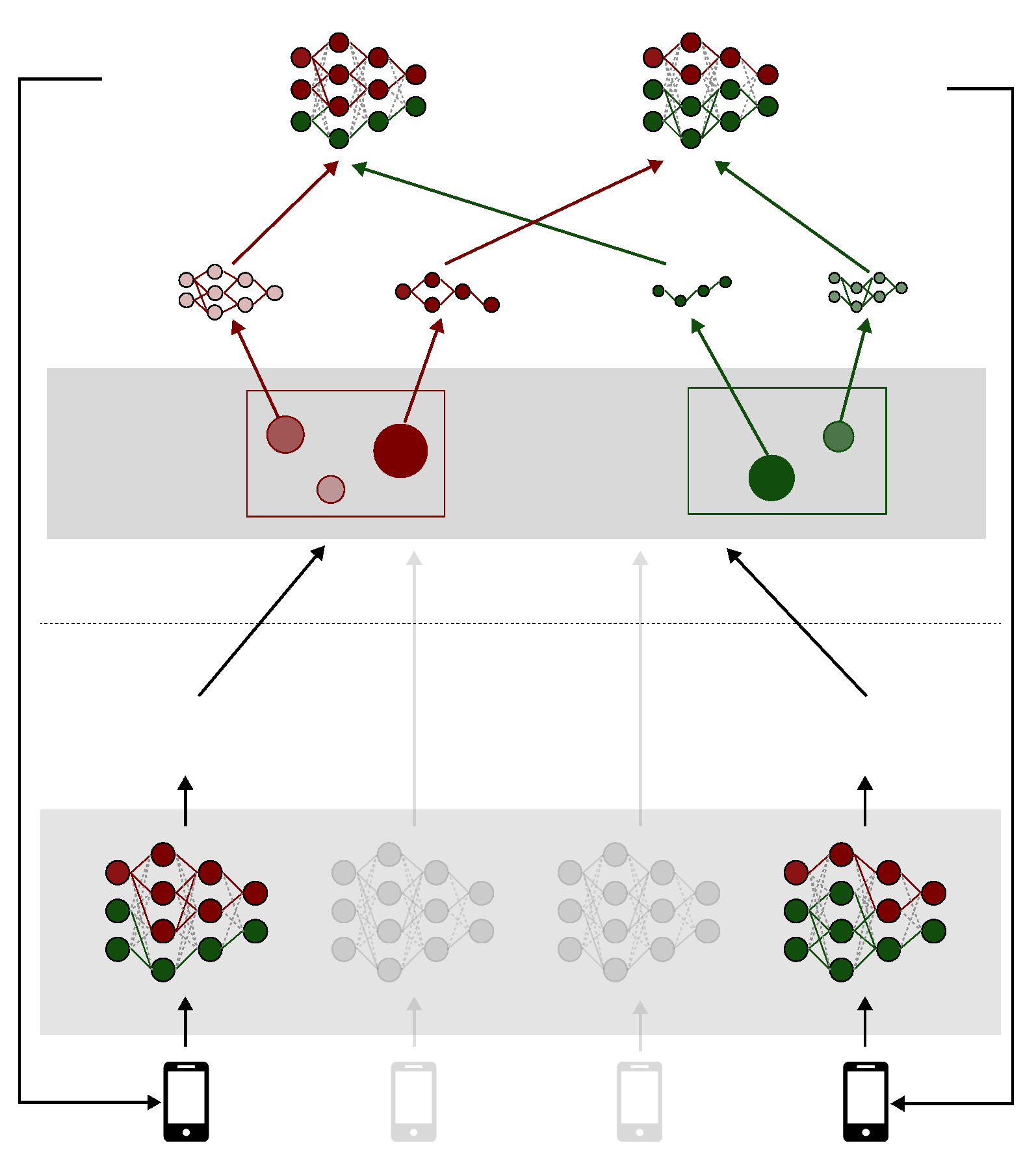}};
        
        \node[text width=3cm] at (2.95, 0.5) {\scriptsize{\textbf{U1}}};
        \node[text width=3cm] at (8.75, 0.5) {\scriptsize{\textbf{U2}}};

        \node[text width=5cm] at (5.1, 1.2) {\small{Local Subnetwork Extraction}};
        \node[text width=3cm] at (1.95, 4.3) {\small{Clients}};
        \node[text width=3cm] at (1.95, 4.8) {\small{Server}};
        \node[text width=3cm] at (1.95, 6.3) {\small{Prototypes\\Clustering}};

        \node[text width=3cm] at (3.4, 1.5) {$\zeta_{1,2}$};
        \node[text width=3cm] at (3.4, 2.6) {$\zeta_{1,1}$};

        \node[text width=3cm] at (9.2, 1.5) {$\zeta_{2,2}$};
        \node[text width=3cm] at (9.2, 2.6) {$\zeta_{2,1}$};

        \node[text width=3cm] at (2, 3.8) {$\qdist{\rho_{1, 1}, \zeta_{1,1}, \omega_{1, 1}}$};
        \node[text width=3cm] at (2, 3.4) {$\qdist{\rho_{1, 2}, \zeta_{1,2}, \omega_{1, 2}}$};

        \node[text width=3cm] at (7.6, 3.8) {$\qdist{\rho_{2, 1}, \zeta_{2,1}, \omega_{2, 1}}$};
        \node[text width=3cm] at (7.6, 3.4) {$\qdist{\rho_{2, 2}, \zeta_{2,2}, \omega_{2, 2}}$};

        \node[text width=3cm] at (3.65, 5.85) {\scriptsize{$C_{1,1}$}};
        \node[text width=3cm] at (4.5, 5.6) {\scriptsize{$C_{1,2}$}};
        \node[text width=3cm] at (4.4, 6.35) {\scriptsize{$C_{1,3}$}};

        \node[text width=3cm] at (7.45, 6.1) {\scriptsize{$C_{2,1}$}};
        \node[text width=3cm] at (8.1, 6.37) {\scriptsize{$C_{2,2}$}};

        \node[text width=3cm] at (2.75, 7) {\small{$\zeta_{C_{1,1}}$}};
        \node[text width=3cm] at (5.35, 7) {\small{$\zeta_{C_{1,3}}$}};

        \node[text width=3cm] at (6.8, 7) {\small{$\zeta_{C_{2,1}}$}};
        \node[text width=3cm] at (9, 7) {\small{$\zeta_{C_{2,2}}$}};

        \node[text width=3cm] at (2.7, 9) {\small{Model\\update\\for U1 $\tilde\theta_1$}};
        \node[text width=3cm] at (8.4, 9) {\small{Model\\update\\for U2 $\tilde\theta_2$}};
        
    \end{tikzpicture}
    \caption{High-level data flow in a FedSub federated learning round. Starting from the botton, each client $u_i$ sends to the server the prototype $\rho_{u, y}$, subnetwork $\zeta_{u,y}$, and classification score $\omega_{u,y}$ for each local class label $y$. The server updates the prototype clusters and generates personalized model updates for each client by fusing the received subnetworks from each cluster.}
    \label{fig:architecture}
\end{figure}

\section{Methodology}
\label{sec:methodology}

Figure~\ref{fig:architecture} illustrates the high-level data flow of a single federated round in FedSub for a binary classification task, highlighting the main operations performed by two representative clients (\textbf{U1} and \textbf{U2}).
The round begins with clients locally extracting all the necessary information for the server to identify similarities in data patterns and generate personalized model updates.
Specifically, for each label $y$, the client $u$ sends to the server: (i) the local prototype $\rho_{u,y}$, (ii) the class-specific subnetwork $\zeta_{u,y}$, and (iii) the classification performance $\omega_{u,y}$.
Then, the server clusters the prototypes to identify similarities among the users at the class level and coordinate the distributed learning process.
Finally, the server aggregates the subnetworks available in its clusters, enabling FedSub to generate a personalized and unique model update for each client. In the following subsections we formalize the algorithms executed by both the clients and the server.

\subsection{Client Side}

As introduced earlier, FedSub assigns an active role to the client in the collaborative training process by performing a series of critical tasks outlined in Algorithm~\ref{algo:client}.
In each communication round, the client first trains its current model on the local dataset for $E$ epochs (line 2).
This step also acts as a fine-tuning phase, refining server updates so the model better aligns with the client’s distribution.
In fact, although FedSub is specifically designed to address the non-IID challenge, some clients may still exhibit unique data patterns that diverge from those of the others, even if some similarities exist.

\begin{algorithm}[t]
\caption{FedSub - Client side}
\label{algo:client}
\KwData{Local training epochs $E$}

\ForEach{communication round $t = 0, 1,2,\ldots,T$ }{

    $h(\theta_t) \gets LocalTraining(h(\theta_{t-1}); E)$

    $\mathbf{P} = \{\rho_y = avg(\phi(x)) : x \in D_y \mid \forall y \in D\}$


    \ForEach{label $y$ in $D$}{

        $\omega_{y} \gets r\_score(h(\theta_{t}), y)$

        \ForEach{layer $l$ of the model}{%
            $\mathbf{\overline{r}}_y^{(l)} \gets avg(\Psi(h(\theta_{t}, x), l) \mid \forall x \in D_y$)
            
            $\mathbf{W_y^{(l)}} \gets \mathbf{\overline{r}_y^{(l)}} \mathbf{W_y^{(l)}}$
        }
    }

    $\mathbf{Z} = \{\zeta_y = \{\mathbf{W_y^{(l)}}\} \mid \forall y \in D, \text{$\forall$ layer $l$ in $h(\theta_t)$}\}$

    $\mathbf{\Omega} = \{\omega_{y} \mid \forall y \in D\}$

    \textbf{send} ($\mathbf{P}$, $\mathbf{Z}$, $\mathbf{\Omega}$) to the server
    
    $\tilde \theta_t \gets$ \textbf{receive} new parameters from the server

    $h(\theta_{t+1}) \gets $ update local model with $\tilde \theta_t$
    
}

\end{algorithm}

Then, similar to FedProto~\cite{Tan_Long_LIU_Zhou_Lu_Jiang_Zhang_2022} and ProtoHAR~\cite{10122911}, the client computes a prototype $\rho_y$ for each class label $y$ available in its local dataset.
Specifically, $\rho_y$ is defined as the mean of the feature representations of all samples labeled as $y$:

\begin{equation}
    \rho_y = \frac{1}{D_y} \sum_{x \in D_y} \phi(x),
\end{equation}

where $D_y$ denotes the subset of the client's local dataset belonging to class $y$, and $\phi(x)$ is the feature extraction function applied to the raw sample $x \in D_y$.
This step captures the central tendency of each class, offering a stable abstraction of the local distribution.
As shown in Algorithm~\ref{algo:client}, the set of all prototypes computed by the client is denoted by $\mathbf{P}$ and is updated whenever the local dataset changes, in order to preserve the relevance of the class prototypes over time.

After prototypes calculation, the client extracts class-level subnetworks, identifying the most relevant components of its local model for each class (lines 4-8).
Specifically, for each class label $y$, the client first calculates a score $\omega_y$ that estimates the reliability of its local model in classifying data samples of $y$ (line 5), which will be later used by the server to weight the client contribution during the subnetworks fusion.
The function to calculate the score, $r\_score(h(\theta), y)$, can be adapted to specific application scenarios and requirements. For instance, it can be defined as the cardinality of the training set for the considered class label, $\omega_y = |D_y|$, thereby prioritizing clients with larger datasets, which may indicate more reliable subnetworks.
Alternatively, it could be set as $accuracy(y)$ to emphasize clients that achieve higher accuracy for that class. Moreover, a combination of these and other metrics can also be used to balance dataset size and model performance when estimating the client reliability during the subnetworks fusion.

The actual extraction of the class-level subnetworks consists of the following three main steps:

\begin{enumerate}

    \item \textbf{Selection of layer-wise relevant units}: for each input sample $x$ labeled with $y$, a forward pass is performed to record neuron activations at each layer.
    Specifically, the function $\Psi(h(\theta_t, x), l)$ extracts the most relevant parameters in layer $l$, providing insights into how the local model processes class-specific information.
    
    \item \textbf{Class-level Activations}:
    since samples of the same class $y$ may activate different model components, the client computes an average activation matrix for each class. This captures common patterns and forms a representative abstraction of class $y$ at each layer.
    The most relevant neural units are identified and stored in the binary mask $\mathbf{\overline{r}}_y^{(l)}$ (line 7), which is then applied to the layer's weights $\mathbf{W_y^{(l)}}$ to obtain a sparse parameters matrix of $l$, zeroing out non-relevant parameters (line 8).
    
    \item \textbf{Subnetwork generation}: the subnetwork corresponding to a specific class label $y$, denoted as $\zeta_{y}$, is thus formed by aggregating the relevant components $\mathbf{W_y^{(l)}}$ across all model layers, encapsulating the essential model pathways involved in recognizing instances of class $y$.

\end{enumerate}

The complete set of extracted subnetworks is represented by $\mathbf{Z}$ (line 9), while the set of recorded class-level performances is indicated with $\mathbf{\Omega}$ (line 10).
Subsequently, the client $i$ transmits its class-specific artifacts for all its classes $\qdist{\rho_{i,y}, \zeta_{i,y}, \omega_{i,y}}$ to the central server (line 11), where they are properly processed to generate a novel set of model parameters $\tilde \theta_t$ specifically tailored to each client’s needs (line 12).

It is important to note that $\tilde{\theta}_t$ results from the fusion (on the server) of multiple subnetworks and, therefore, it may be sparse, not representing a complete update of the entire model.
For this reason, the client updates the local model by integrating only the novel parameters from $\tilde{\theta}_t$ (line 13).
The updated model, $h(\theta_{t+1})$, will then undergo additional fine-tuning in the next communication round to further adapt it to the peculiarities of the local data.

Subnetworks extraction represents the most critical part of the client algorithm, strictly depending on the accuracy with which the most relevant model components are selected, while redundant or irrelevant parameters are discarded.
To achieve the best accuracy, we explore two alternative approaches for the definition of $\Psi(h(\theta_t, x), l)$, namely (i) \textbf{Na\"ive Activation-Based Selection}, and (ii) \textbf{LRP-Based Subnetwork Selection}.
In the former method, the client selects the parameters associated with neurons that exhibit positive activation during the forward pass.
Therefore, the function $\Psi_{Naive}$ extracts both weights and biases for activated neurons, while setting others to zero:

\begin{equation}
    \Psi_{Naive} = \left\{
        \begin{array}{lr}
            [w_j, b_j], & \text{if } \sigma_j \left( \sum_{i=1}^n w_i x_i + b \right) > 0\\
            0, & \text{otherwise}\\
        \end{array}
    \right\}_{j=1}^{\mathcal{U}_l}
\end{equation}

where $\mathcal{U}_l$ denotes the number of neural units in the layer $l$, $\sigma_j$ is the activation function of neuron $j \in \mathcal{U}_l$, $x_i$ denotes one of its $n$ inputs, $w_i$ its associated weight and $b_j$ the bias term.

This approach is both computationally simple and interpretable, as a positive activation indicates an active contribution to data processing.
However, it is sub-optimal since it may retain redundant parameters, unnecessarily increasing the subnetwork size.
Therefore, we also explore an alternative approach based on \emph{Layer-wise Relevance Propagation} (\emph{LRP})~\cite{10.1371/journal.pone.0130140} to identify the most relevant components of the subnetworks, which can also potentially reduce the subnetwork size.
LRP is an Explainable-AI technique that has been applied in various fields, including general image recognition, medical imaging, and natural language processing~\cite{Montavon2019}.
More recently, it has also been used as a fundamental technique to guide model pruning for neural network optimization~\cite{YEOM2021107899}, but it has never been applied in the context of FL.

LRP propagates the classification score backward through the network, assigning relevance scores to inputs and revealing which features contribute most to a prediction.
LRP follows a general conservation principle, ensuring that the total relevance at each layer is preserved as it propagates backward: if a neuron in a later layer receives a certain amount of relevance, it distributes that score to its contributing neurons in the previous layer. Formally, this can be expressed as follows:

\begin{equation}
    R_j = \sum_k \frac{z_{jk}}{\sum_{j} z_{jk}} R_k,
\end{equation}

where $R_k$ is the relevance of neuron $k$ in the upper layer, $R_j$ indicates the relevance of neuron $j$ in the lower layer, and $z_{jk}$ is the contribution distributed from neuron $j$ to neuron $k$. 

We follow the LRP approach by implementing the subnetwork extraction process as the composition of 3 main steps: (i) a standard forward pass is performed to collect the units activations at each layer; (ii) the final classification score is backpropagated through the network; (iii) only the parameters of the neurons with a positive relevance score are included in the final subnetwork.
Formally, the LRP-based subnetwork extraction can be formalized as follows:

\begin{equation}
    \Psi_{LRP} = \left\{
        \begin{array}{lr}
            [w_j, b_j], & \text{if } R_j > 0\\
            0, & \text{otherwise}\\
        \end{array}
    \right\}_{j=1}^{\mathcal{U}_l}
\end{equation}

where $R_j$ indicates the relevance score assigned to the neural unit $j$ in layer $l$.

Within the LRP framework, the backpropagation of relevance scores is guided by a propagation rule.
Based on the type of layer (e.g., fully-connected, convolutional, or pooling) different rules have been proposed in the literature.
In FedSub, we specifically adopt relevance quantities computed with $\text{\emph{LRP-}}\alpha\beta$ rule as a criterion to extract the class-level subnetworks, since it has been especially developed for feedforward deep neural networks with ReLU activations and assumes positive logit activations (i.e., the model output) for decomposition~\cite{10.1371/journal.pone.0130140}.
Moreover, this rule distinguishes positive and negative contributions during backpropagation, tuning their influence through parameters $\alpha$ and $\beta$:

\begin{equation}
    R_j = \sum_k \left( \alpha \frac{z_{jk}^{+}}{z_{k}^{+}} - \beta \frac{z_{jk}^{-}}{z_{k}^{-}} \right)R_k,
\end{equation}

where $(\cdot)^{+}$ and $(\cdot)^{-}$ indicate respectively positive and negative contributions from the neural unit $k$ of the upper layer towards the neuron $j$ in the lower layer, and $\alpha$ and $\beta$ are parameters for tuning the focus on excitatory (i.e., positive) or inhibitory (i.e., negative) influences in the model.

In the original $LRP-\alpha\beta$ rule, the parameters are set to $\alpha=1$ and $\beta=0$, considering only positive contributions and strictly following the conservation principle~\cite{10.1371/journal.pone.0130140}.
Later works showed that relaxing this constraint, for example setting $\alpha=2$ and $\beta=1$, can enhance interpretability by also accounting for negative contributions~\cite{samek2016interpreting, Montavon2019}.
In this paper, we test both rules, comparing them with the \emph{Na\"ive} approach with the aim of extracting the most relevant class-level subnetworks for the federated learning task.

In terms of computational complexity, the calculation of class prototypes, when an update is required, involves simple feature-wise averaging over local samples. This operation has time complexity $\mathcal{O}(|D| \cdot d)$, where $|D|$ represents the total number of samples in the local dataset, and $d$ is the feature dimensionality.
On the other hand, the extraction of label-specific subnetworks can be performed using different strategies. In the \emph{Na\"ive} approach, subnetworks are selected based on neuron activations obtained through a single forward pass, resulting in a time complexity of $\mathcal{O}(|D| \cdot |w|)$ where $|w|$ represents the total number of parameters in the local model.
In contrast, the LRP-based methods require both a forward pass and a backward relevance propagation step, which has a computational cost comparable to a traditional gradient backpropagation~\cite{Montavon2019}. Thus, the overall time complexity of the LRP-based approaches can be approximated to $\mathcal{O}(|D| \cdot 2 \cdot |w|)$.

It is worth noting that, since subnetworks are extracted independently for each label in $D$, and modern neural models can process multiple inputs in parallel via batch processing, the overall subnetwork extraction procedure is highly parallelizable.
Thus, the effective complexity mainly depends on the number of model parameters rather than on the dataset size or number of classes, making the method feasible even on resource-constrained devices.

Finally, it is important to acknowledge that in scenarios involving an extremely large number of classes, such as image recognition tasks with thousands of categories, the number of extracted subnetworks may increase substantially.
In this case, although FedSub remains highly parallelizable and computationally manageable, the proliferation of subnetworks could raise memory and communication demands, particularly in bandwidth-limited environments, thus becoming a potential scalability bottleneck.
By contrast, the application scenarios considered in this work, including HAR and mHealth, are typically characterized by a relatively small number of classes, where such scalability concerns do not arise, and thus FedSub remains an efficient and viable solution.

\subsection{Server side}

As described in Algorithm~\ref{algo:FedSub_server}, the server begins each round by randomly selecting $m$ clients from the active pool $\mathbf{U}$ (line 3). The selected clients then execute their local procedures and return the class-level artifacts (lines 4–6) that form the basis for the server-side coordination mechanism.

A core component of this mechanism is the artifacts cache (represented as $\mathbf{P}, \mathbf{Z}, \mathbf{\Omega}$) maintained by the server to store the most recent data shared by the clients.
Specifically, at each round, only the entries corresponding to the participating clients are updated, while clustering is performed over all the cached prototypes.
This strategy allows the server to exploit a richer and more diverse set of representations accumulated over time, rather than relying solely on the limited view provided by the currently active clients.
As a result, it reduces sensitivity to sampling variability and mitigates the risk of unstable or noisy clusters when only a small or unbalanced subset of clients participates.

\begin{algorithm}[t]
\caption{FedSub - Server side}
\label{algo:FedSub_server}
\KwData{Communication rounds $T$, total user clients $\mathbb{U}$ , number of clients at each round $m$.}

$\mathbf{P}, \mathbf{Z}, \mathbf{\Omega} \gets$ init artifacts cache

\ForEach{communication round $t = 0,1,2,\ldots,T$}{%
      $\mathbb{U}_m \gets $ randomly select $m$ clients from $\mathbb{U}$ 

      \ForEach{client $u \in \mathbb{U}_m$ in parallel}{
        \textbf{execute} $u$ code
      }

      Update $\mathbf{P}, \mathbf{Z},\mathbf{\Omega}$ with data \textbf{received} by $\mathbb{U}_m$

      
      $\mathbf{C_p} \gets$ ClusteringPrototypes($\mathbf{P}$)

      \ForEach{client $u \in \mathbb{C}_m$ in parallel}{%
        $\mathbf{Z}_u^{'} \gets $ SubnetworksFusion($u$, $\mathbf{C_p}$, $\mathbf{Z}$)
        
        $\tilde{\theta}_{u,t} \gets $ CombineSubnetworks($\mathbf{Z}_u^{'}$)
        
        \textbf{send} $\tilde{\theta}{u,t}$ to $u$
      }
}
\end{algorithm}

The server then performs a class-level clustering of the prototypes available in $\mathbf{P}$ to infer similarities in the clients' underlying data distributions (line 7).
To this aim, we employ the well-known K-means algorithm, dynamically determining the optimal number of clusters at each round and for each class-level clustering.
Specifically, every round a range of candidate values is evaluated, starting from a minimum of 2 clusters up to the current number of available prototypes, and the value of K that minimizes the Davies–Bouldin Index (DBI)~\cite{4766909} is selected.
This dynamic selection is crucial given the highly evolving nature of the target environment: as new data distributions emerge or existing ones shift, the underlying cluster configuration may change accordingly. By re-estimating the optimal K at every round, we ensure that the clustering process remains aligned with these changes, allowing the model to adapt to varying prototype over time.
While alternative clustering methods could also be adopted (e.g., Hierarchical Clustering (HC)~\cite{10.1145/3321386} or DBSCAN~\cite{6814687}), we selected K-means because of its lower computational complexity: unlike HC and DBSCAN, which may reach quadratic or cubic complexity in the worst case~\cite{10.1145/3068335, Nielsen2016}, K-means is generally more efficient and thus suitable for scalable deployments.

Once clusters are formed, the server generates personalized updates in three steps: (i) it fuses the subnetworks within each cluster that client $u$ belongs to (line 9); (ii) it combines these subnetworks into a personalized update $\tilde{\theta}_{u,t}$ (line 10); and (iii) it sends $\tilde{\theta}_{u,t}$ back to $u$ for local integration (line 11).
Subnetworks fusion is the server’s most critical and innovative operation, with a major impact on collaborative learning.
The goal of a fusion strategy is to create class-level model updates that integrate knowledge contributions from users with similar behaviors (i.e., belonging to the same cluster) while also accounting for diversity in data and model activations.
In this work, we explore the three strategies:

\subsubsection{\textbf{Cluster AVG}}

This strategy extends FedAvg to the cluster level, where all members contribute to the subnetwork fusion with weights proportional to their reliability scores.
Specifically, the subnetworks within a cluster are averaged to produce an aggregated subnetwork, with each member’s contribution weighted by their shared reliability scores $\omega_{u,y}$.
Therefore, each element of the average parameter matrix $\mathbf{W^{(l)}_{\overline{C}}}$ of the layer $l$ for the cluster $C$ is formally defined as follows:

\begin{equation}
    \label{eq:cluster_avg}
     \mathbf{W^{(l)}_{\overline{C}}}[i, j] =
     \frac{
        \sum_{u \in C} \mathbf{W^{(l)}_{u,t}}[i,j] \cdot \omega_{u, y}}
        {\sum_{u\in C} \omega_{u,y}},
\end{equation}

where $\{u_1, u_2, \ldots, u_n\}$ is the set of clients grouped in $C$, $y$ is the label associated with the cluster, while $\omega_{u,y}$ and $\mathbf{W^{(l)}_{u,t}}$ indicate respectively the reliability score and parameters matrix for the layer $l$ of the subnetwork shared by the client $u$ at round $t$ for the class label $y$.

Finally, the subnetwork update for the client $u$ and layer $l$, $\mathbf{W^{(l)}_{u,t+1}}$, is calculated by combining the different $\mathbf{W^{(l)}_{\overline{C}}}$ of each cluster in which $u$ is a member, as follows:

\begin{equation}
    \mathbf{W^{(l)}_{u, t+1}}[i, j] = avg(\{\mathbf{W^{(l)}_{\overline{C}}}[i,j] \mid \forall c \in \mathbf{C^u}\}),
\end{equation}

where $\mathbf{C^u}$ indicates the set of clusters $u$ belongs to, and $avg(\cdot)$ denotes the element-wise average among a set of parameter matrices.

\subsubsection{\textbf{Cluster Leadership}}

The key idea behind this strategy is that the best-performing models should steer the collaborative learning process.
Here, the highest-performing client in each cluster, the \emph{cluster leader}, represents the entire group.
All members then update their subnetworks by adopting the leader’s parameters.

Formally, the parameter matrix for client $u$ at layer $l$ is defined as follows:

\begin{equation}
    \mathbf{W^{(l)}_{u, t+1}}[i, j] = avg(\{\mathbf{W^{(l)}_{c^*}}[i,j] \mid \forall c \in \mathbf{C^u}\}),
\end{equation}

where $c^*$ represents the leader of the cluster $c$, determined as $c^*=argmax(\{\omega_{k,y} \mid \forall k \in C\})$, while $avg(\cdot)$ denotes the element-wise average that integrates the subnetworks of the cluster leaders across all clusters to which $u$ belongs.

\subsubsection{\textbf{Overlapping Components}}

This is the most conservative strategy, updating only the components shared by all subnetworks in a cluster.
This method allows models to benefit from contributions made by similar clients while maintaining the distinctive characteristics of each local model.

More specifically, the subnetworks within a cluster are combined using a modified version of Equation~\ref{eq:cluster_avg} that considers only relevant components across all clients (i.e., non-zero matrix elements), as described in the following formula:

\begin{equation}
    \small
    \mathbf{W^{(l)}_{\overline{C}}}[i,j] =
    \begin{cases}
        \frac{\sum_{u \in C} \mathbf{W^{(l)}_{u, t}}[i,j] \cdot \omega_{u,y}}{\sum \omega_{u,y}}, & if~\mathbf{W^{(l)}_{u,t}}[i,j] \neq 0, \forall u \in C\\
        0,  & otherwise.
  \end{cases}
\end{equation}

In other words, if a given neural unit is deemed relevant in all the subnetworks within the cluster $C$, then $\mathbf{W^{(l)}_{\overline{C}}}$ will contain its associated parameters, averaged across all the subnetworks. Otherwise, the unit will be considered non-relevant, and the related parameters in $\mathbf{W^{(l)}_{\overline{C}}}$will be set to 0.

Following this, the final update for each client is derived by adding the average contributions from all clusters to which the client belongs, still focusing only on the overlapping components among all the subnetworks.
Formally, this can be expressed as follows:

\begin{equation}
    \mathbf{W^{(l)}_{u, t+1}}[i,j]=
    \begin{cases}
        \scriptstyle
        avg(\mathbf{W_{\overline{C}}^{(l)}}[i,j]), \forall C \in \mathbf{C^u}\\
        \scriptstyle if~\mathbf{W^{(l)}_{u,t}}[i,j] > 0 ~\wedge \mathbf{W_{\overline{C}}^{(l)}}[i,j] > 0, \forall C \in \mathbf{C^u} \\\\
        \scriptstyle
        \mathbf{W^{(l)}_{u, t}}[i,j], otherwise,
  \end{cases}
\end{equation}

where each element of the resulting parameter matrix for the client $u$ will be replaced by the average of the corresponding elements in $\mathbf{W^{(l)}_{\overline{C}}}$ of each cluster $C$ that contains $u$, but only if they share the same relevant neurons. Otherwise, client $u$ will retain its original parameters.

Through these operations, the FedSub server coordinates the learning process and synthesizes updates tailored to each client’s characteristics.
Moreover, by clustering similar clients, and carefully aggregating subnetworks, the server ensures that the FL system remains robust and capable of handling diverse and distributed datasets.

Regarding computational complexity, the most intensive operation performed by the server is the prototypes clustering. For each label, the execution time of K-means can be approximated to $\mathcal{O}(I \cdot K \cdot \mathbb{\varrho}_y \cdot d)$, where $I$ is the number of iterations, $K$ the number of clusters, $\mathbb{\varrho}_y$ the number of prototypes for the label $y$, and $d$ is their dimensionality~\cite{7065640}.
In addition, consider that this step remains particularly scalable, due to its label-wise separability, which enables parallel processing across labels and significantly reduces runtime.
Once the clustering configuration has been determined, the fusion of subnetworks is executed in parallel for each client. Moreover, since the linear operations included in the fusion strategies are defined on a per-layer basis, this step can be further parallelized across model layers.
Therefore, given this high degree of parallelism across labels, layers, and clients, FedSub scales efficiently with both the number of clients and the number of classes, while keeping the computational load distributed across the system.

\begin{table*}[t]
\caption{Characteristics of the simulated application scenarios and corresponding datasets.}
\label{tab:datasets}
\centering
\begin{tabular}{lllp{20mm}rrrrrr}
\toprule
\textbf{Scenario} & \textbf{Dataset} & \textbf{Device Type} & \textbf{Modalities} & \textbf{Users} & \textbf{Samples} & \textbf{Features} & \textbf{Classes} & \textbf{Min Class} & \textbf{Major Class} \\
\midrule
HAR & WISDM~\cite{10.1145/1964897.1964918} & Smartphone & Acc. & 36 & 53,776 & 3 & 6 & 0.08 & 0.42 \\
Stress Detection & WESAD~\cite{10.1145/3242969.3242985} & Wearable & ECG, EDA & 15 & 6,189 & 30 & 4 & 0.12 & 0.40\\
Sleep Detection & DREAMT~\cite{pmlr-v248-wang24a} & Wearable & Acc., Temp., HR, EDA, BVP & 100 & 189,211 & 8 & 2 & 0.03 & 0.42\\
\bottomrule
\end{tabular}
\end{table*}

\subsection{Privacy Considerations}

\begin{figure}[t]
     \centering
     \includegraphics[width=0.99\columnwidth]{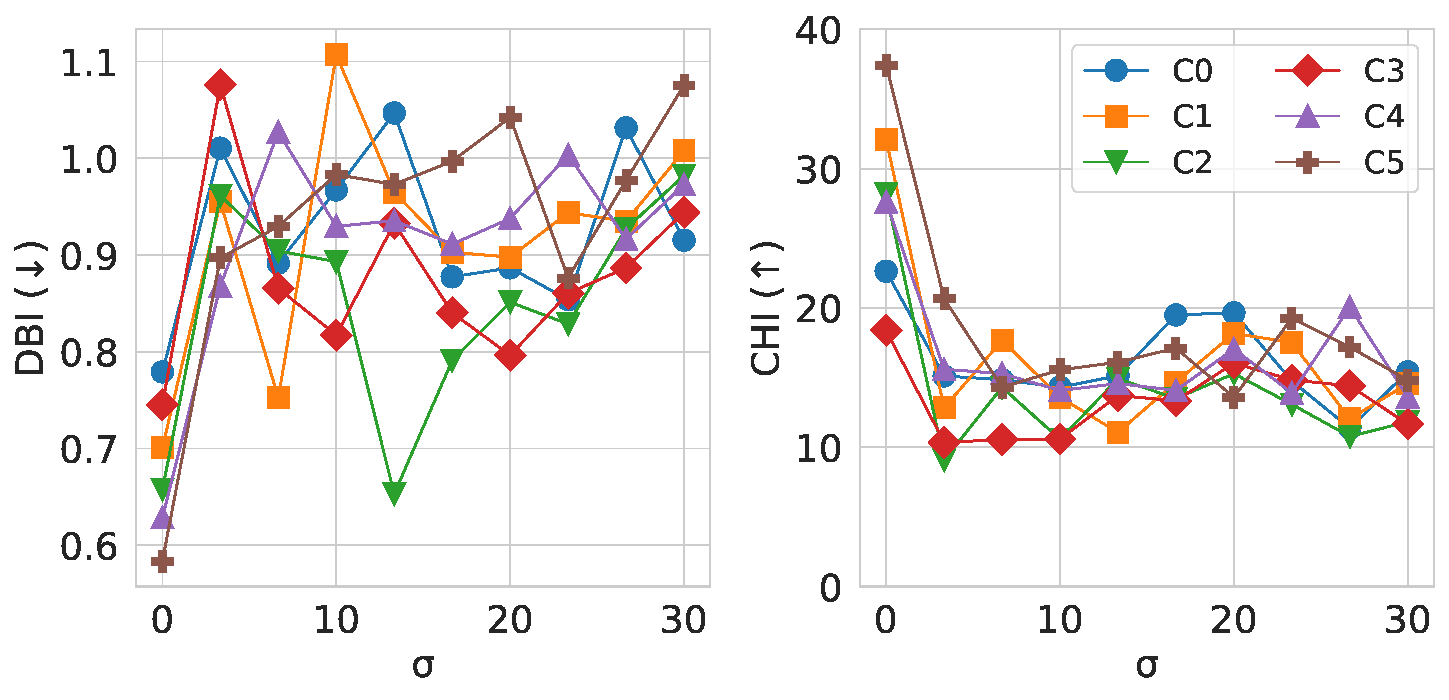}
     \caption{Quality of prototype clustering in terms of DBI and CHI, under varying levels of perturbation.}
     \label{fig:dp_clustering}
\end{figure}

Although prototypes are common in some PFL approaches, they may inadvertently reveal sensitive information to the server.
To address this risk in applications requiring stronger privacy guarantees, mechanisms such as Differential Privacy (DP)~\cite{10.1145/3490237} can be seamlessly integrated into FedSub to enhance the anonymization of shared prototypes. In practice, clients can inject calibrated random noise into each prototype, preventing the server from reconstructing or inferring the underlying data. This noise is typically drawn from a Gaussian or Laplacian distribution and scaled according to a privacy budget, which must be carefully tuned to strike a balance between privacy protection and the accuracy of server-side clustering.

As an illustrative example, Figure~\ref{fig:dp_clustering} reports prototype clustering performance at the class level using k-means with $k=5$ on the WISDM dataset~\cite{10.1145/1964897.1964918}, which contains tri-axial accelerometer data across six human activities. The level of client-side noise (measured as its standard deviation, $\sigma$) is systematically varied to study its impact on the clustering configuration.
Since there is no ground-truth cluster labels, clustering quality is assessed with standard unsupervised metrics: DBI (lower is better) and the Calinski–Harabasz Index (CHI, higher is better). As expected, higher levels of prototype distortion progressively degrade clustering quality.

It is worth noting that, in this illustrative experiment, we fixed the number of clusters solely to isolate and highlight the effect of noise on K-means.
In contrast, FedSub dynamically selects the value of $K$ at each round, which makes it more challenging to determine the optimal trade-off between the noise level and the resulting clustering configuration.
While clustering metrics can serve as a direct proxy for personalization quality, a comprehensive end-to-end evaluation of FedSub under varying noise conditions would offer additional empirical insights.
Such a dedicated analysis, however, falls beyond the scope of this paper and will be addressed in future work.
In the same direction is the exploration of clustering algorithms that incorporate DP by design. For instance, the modified k-means algorithm proposed by Xia et al.~\cite{XIA2020101699} may offer a more principled approach to achieve favorable trade-offs between privacy and utility, even when operating under strong perturbations. Currently, there is no available implementation of this proposal, but we are planning to integrate it in future evaluations. 

\section{Experimental Evaluation}
\label{sec:experiments}

In this section, we evaluate FedSub through a series of experiments on three real-world datasets derived from mHealth and remote monitoring applications. Specifically, they cover three use cases: HAR from inertial sensors, and Stress and Sleep Detection from physiological signals collected by wearables.
In the following, we provide a detailed description of the selected datasets and preprocessing steps. Then, we compare FedSub performance with the most prominent PFL algorithms under both static and dynamic data generation settings.

\subsection{Datasets and application scenarios}

Table~\ref{tab:datasets} offers a summary of the main characteristics of each scenario, highlighting the specific sensor modalities and features considered in our simulations.

\textbf{Human Activity Recognition (HAR)}:
In this scenario we refer to  \emph{Wireless Sensor Data Mining} (\emph{WISDM})~\cite{10.1145/1964897.1964918} as HAR dataset.
It contains 3-axial accelerometer (Acc.) data collected from smartphones of 36 subjects, presenting significant variability in movement patterns across individuals. The dataset is labeled with six activity classes: \verb|Walking|, \verb|Jogging|, \verb|Upstairs|, \verb|Downstairs|, \verb|Sitting|, and \verb|Standing|.

We applied a 1-second sliding window without overlap to the smartphone accelerometer data, averaging the three components of the accelerometer readings. This preprocessing step resulted in a standardized dataset of 53,776 samples, with each user contributing an average of 1,493.78 samples.

\textbf{Stress Detection}:
The second scenario exploits the \emph{Wearable Stress and Affect Detection} (\emph{WESAD}) dataset~\cite{10.1145/3242969.3242985}.
It derives from a clinical protocol designed to distinguish among 4 different stress conditions based on the users' physiological response to external stimuli. The dataset includes data from 15 subjects, collecting physiological signals such as electrocardiogram (ECG) and electrodermal activity (EDA), labeled with four distinct states: \verb|Baseline|, \verb|Stress|, \verb|Amusement|, and \verb|Meditation|.
We focussed on ECG and EDA signals from a chest-worn device, preprocessed by applying a sliding window of $10,000$ samples with 50\% overlap.
Each window has been processed through the NeuroKit2 toolkit~\cite{Makowski2021} to extract 30 features commonly used for this task, including time- and frequency-domain measures to model the ECG signal~\cite{Singh2023}, and standard statistics related to both phasic and tonic changes in the EDA signal~\cite{10024755}, producing a final dataset of 6,189 samples, with an average of 412.6 samples per user.

\textbf{Sleep Detection}:
The third and last scenario is based on the \emph{Dataset for Real-time Sleep Stage Estimation using Multisensor Wearable Technology} (\emph{DREAMT})~\cite{pmlr-v248-wang24a}.
DREAMT contains data from 100 participants, including high-resolution physiological signals captured by the Empatica E4 wristband, with labels annotated by expert sleep technicians.
Similarly to the original dataset publication, we perform the classification task between \verb|Sleep| and \verb|Wake| states, by using the following 8 features of the physiological data: blood volume pulse (BVP), inter-beat interval (IBI), EDA, skin temperature (Temp.), heart rate (HR), and tri-axial accelerometer data.
After feature standardization, the final dataset includes a total of 189,211 samples, with each user contributing an average of 1,891.12 samples.

\subsection{Reference baselines}

Among the related works in Section~\ref{sec:related}, we selected five baseline algorithms:

\begin{itemize}
    \item \textbf{FedAvg}~\cite{pmlr-v54-mcmahan17a}: The de-facto standard in FL without personalization, where client models are simply averaged to produce a single global model.

    \item \textbf{FedHome}~\cite{9296274}:
    it applies data augmentation for personalization. After training a global model with FedAvg, each client locally fine-tunes using SMOTE to generate a balanced dataset with synthetic samples.

    \item \textbf{FedCLAR}~\cite{9762352}:
    a clustering-based method that creates different global models. Once the FedAvg model stabilizes, clients are clustered by model similarity, and a separate model is computed for each cluster. A final local fine-tuning round adapts the cluster model to individual clients.

    \item \textbf{ProtoHAR}~\cite{10122911}: a prototype-based PFL approach where clients share class-wise prototypes with the server, which aggregates them to form global prototypes. These are then used by clients to regularize local training and mitigate the effects of non-IID datasets.

    \item \textbf{Ditto}~\cite{pmlr-v139-li21h}:
    a regularization-based method where a proximal penalty constrains local updates to remain close to the global model.

\end{itemize}

This selection allows us to compare FedSub with the PFL state-of-the-art, including those algorithms that are conceptually closer to FedSub. We also considered ProtoHAR as an implicit baseline for those models that it already overtakes in previous experiments (i.e., MOON)~\cite{10122911}.
Moreover, to perform a fair comparison within the same simulation environment, we extended the evaluation framework provided by the authors of ProtoHAR~\footnote{\url{https://github.com/cheng-haha/ProtoHAR}}, implementing FedSub, FedCLAR, and Ditto~\footnote{We ported the code available in the official Ditto repository (\url{https://github.com/litian96/ditto}) from Tensorflow 1.10 to PyTorch 2.5}, while utilizing existing implementations of FedAVG, FedHome, and ProtoHAR already available within the original framework.

\begin{figure*}[t]
     \centering
     \begin{subfigure}[t]{0.32\textwidth}
         \centering
         \includegraphics[width=\textwidth]{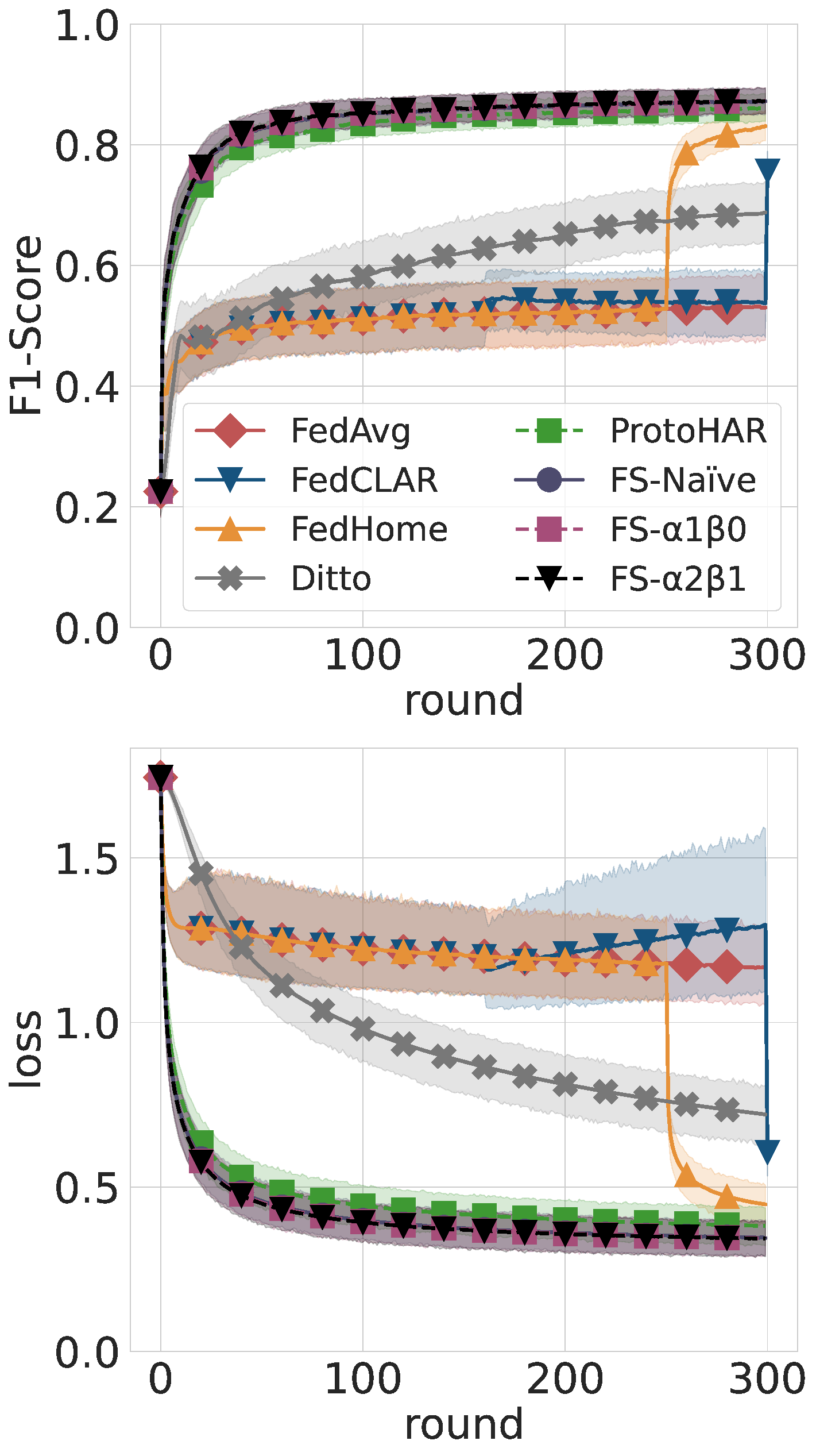}
         \caption{HAR}
         \label{fig:static_wisdm}
     \end{subfigure}
     \begin{subfigure}[t]{0.32\textwidth}
         \centering
         \includegraphics[width=\textwidth]{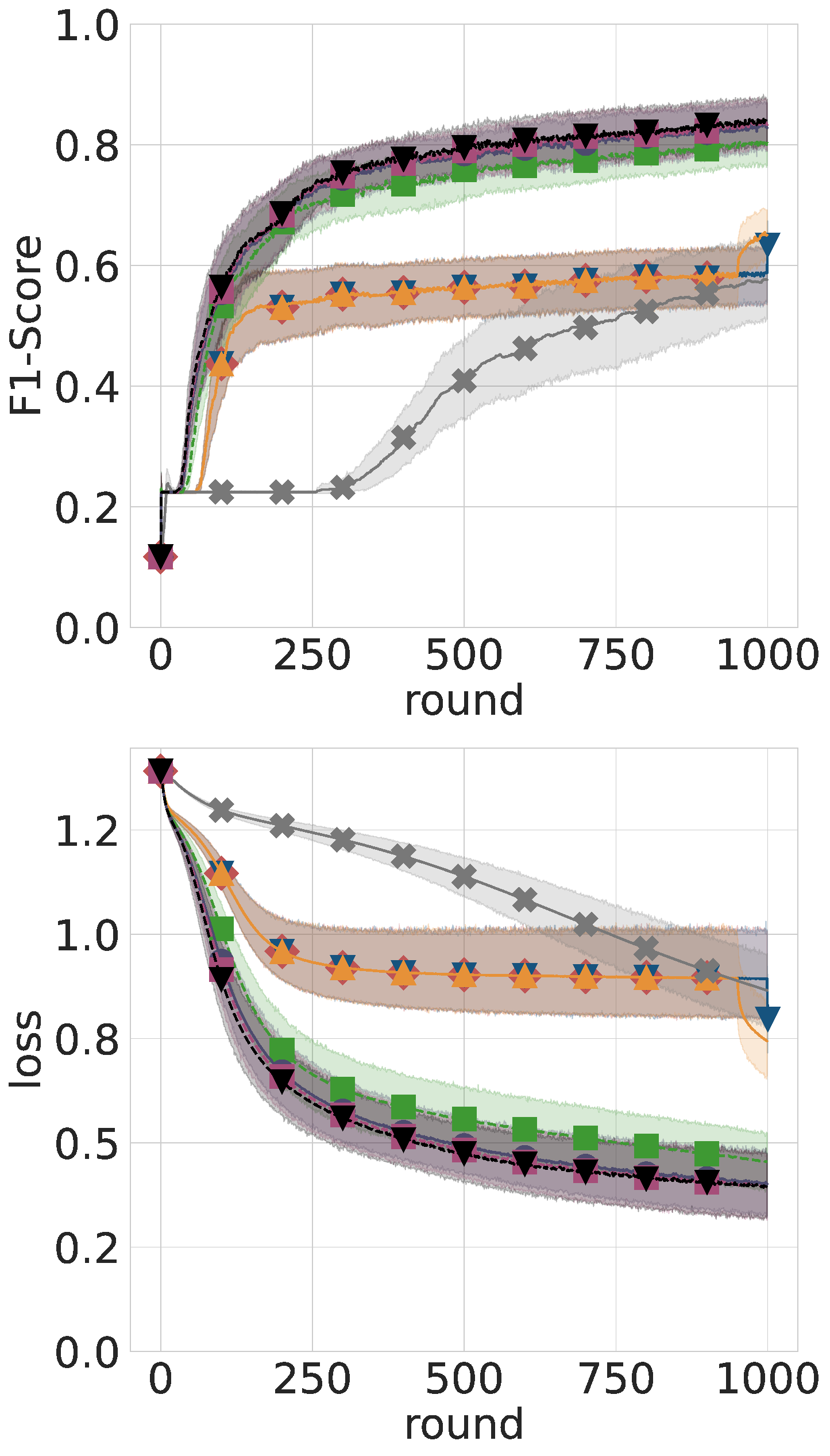}
         \caption{Stress Detection}
         \label{fig:static_wesad}
     \end{subfigure}
     \begin{subfigure}[t]{0.32\textwidth}
         \centering
         \includegraphics[width=\textwidth]{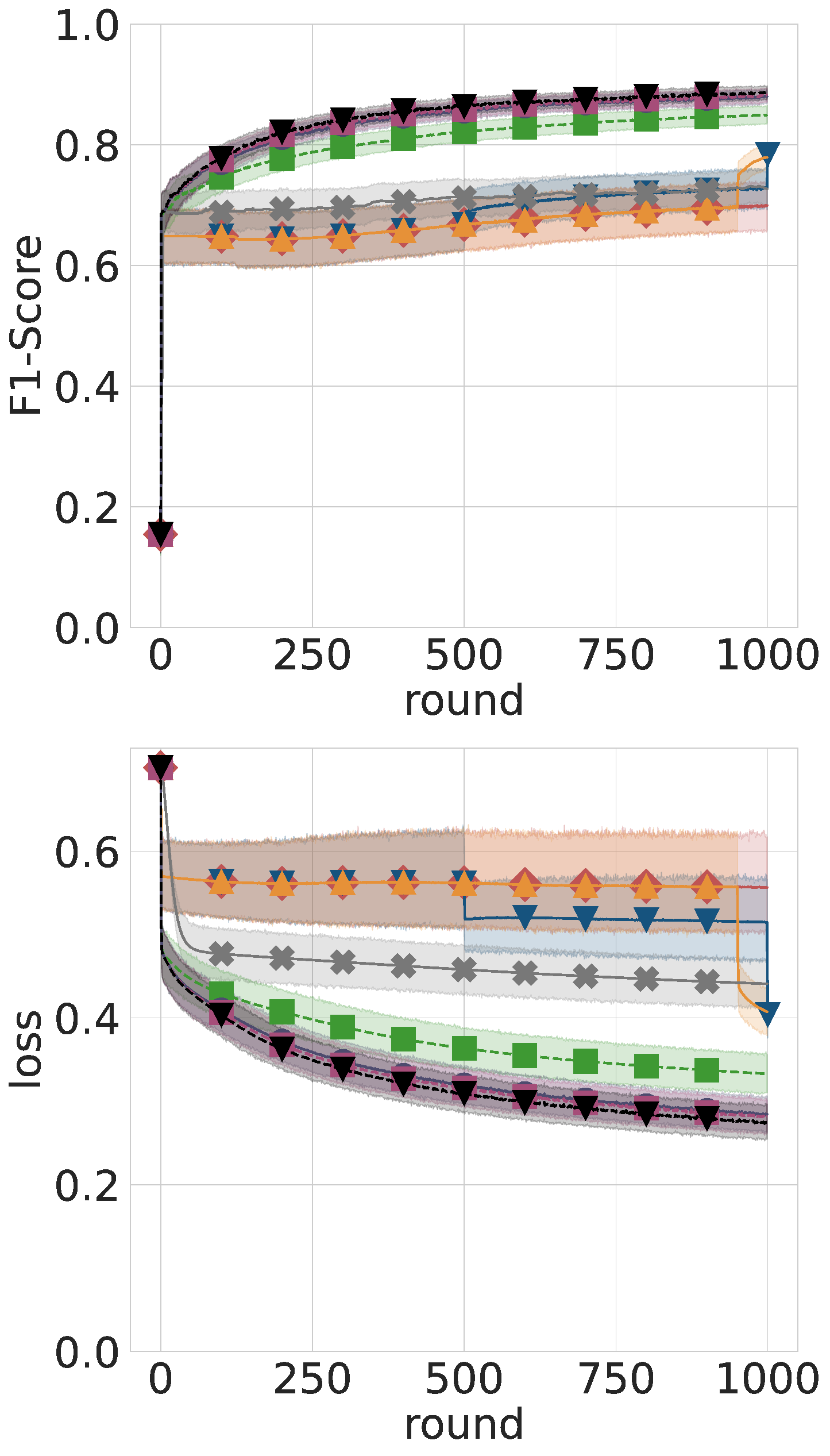}
         \caption{Sleep Detection}
         \label{fig:static_dreamt}
     \end{subfigure}
    \caption{Average F1-Score and loss value with 95\% confidence intervals in the static scenario.}
    \label{fig:static_scenario}
\end{figure*}

\subsection{Experimental Setup}

As far as the classification model is concerned, we selected a feedforward architecture, consisting of two hidden layers with 128 and 512 neurons, respectively. The input and output layers have been configured according to the dataset used: (i) for WISDM, we used 3 input neurons for the accelerometer data and 6 output neurons for the different activities; (ii) for WESAD, we employed 30 input neurons, corresponding to the extracted ECG and EDA features, and 4 output neurons for the stress and affective states; (iii) and for DREAMT, we used 8 neurons for the multimodal input features and 2 outputs to distinguish between sleep and wake stages.
The Rectified Linear Unit (ReLU) was used as activation function, Cross-Entropy as loss function, and Stochastic Gradient Descent (SGD) as optimizer.

We report results after a sufficient number of rounds to ensure the algorithms convergence in each scenario: 300 rounds for HAR and 1000 rounds for both Stress and Sleep Detection.
In terms of performance metrics, we adopted the
average F1-Score and test loss value across all the simulated clients.
While the former assesses classification performance in scenarios with underrepresented classes (as in our datasets, see Table~\ref{tab:datasets}), the latter provides insights into the model ability to generalize to new data samples,
capturing both predictive accuracy and confidence.
Moreover, in the experiments we considered all the available clients as active members in each round, one epoch of local training for all algorithms, and one fine-tuning epoch for FedHome and FedCLAR.

Regarding parameters selection, for ProtoHAR we adopted the same settings reported in~\cite{10122911}.
As far as FedCLAR is concerned, we used the best values found via grid search: a clustering threshold of $0.0001$, with clustering performed in the middle of the simulation (round 160 for HAR, round 500 for Stress and Sleep Detection), when the global model is expected to be stable. For FedHome, we considered the same configuration of~\cite{10122911}, starting fine-tuning 50 rounds before the end of training.
Finally, as far as Ditto, we empirically tested a range of $\lambda$ values (i.e., $0$, $0.1$, $0.5$, $1$, $5$) to assess its regularization sensitivity. Interestingly, we observed very similar performance across all settings suggesting that, in our experimental setup, the regularization mechanism has a limited effect compared to the main classification loss, possibly because $\lambda$ mainly constrains the deviation from the global model rather than effectively capturing client-specific patterns.
Therefore, for fairness and consistency with prior work, we selected $\lambda = 1$ as the default setting, which is also the most commonly reported value in the literature~\cite{pmlr-v139-li21h}.

For FedSub, we report results with both \emph{Naïve} and \emph{LRP-based} subnetwork extraction.
On the server, we focus on the Overlapping Components strategy, which consistently outperformed the others (see Section~\ref{sec:hyperparams}).
We also performed a grid search for client relevance scores. Specifically, we tested four options: (i) $\omega_y = 1$ (equal contributions); (ii) $\omega_y = |D_y|$ (dataset size); (iii) $\omega_y = accuracy(y)$ (classification accuracy); and (iv) $\omega_y = accuracy(y)|D_y|$ (combination of both).
Scores were normalized to $[0,1]$ within each cluster, by dividing each score by the maximum value in the cluster.
As discussed in Section~\ref{sec:hyperparams}, only the relevance score (ii) caused slower convergence, while the others gave similar performance.
Therefore, to fairly assess FedSub itself, we report results with the simplest setting, i.e., equal contributions and \emph{Overlapping Components} fusion.

\subsection{Static scenario}

In this scenario, each user corresponds to a distinct FL client with full access to its dataset from the start.
Specifically, each client's data is first partitioned in training ($70\%$) and test ($30\%$), using stratified sampling to ensure adequate class representation in both subsets.
Moreover, in this setting, both sets remain unchanged throughout the whole simulation, as well as the prototypes computed by FedSub.
The results shown in Figure~\ref{fig:static_scenario} clearly distinguish two groups of algorithms: the former is represented by FedAVG-like solutions, with FedCLAR, FedHome, and Ditto; while the latter, which includes ProtoHAR and FedSub, achieving greater personalization leveraging on prototypes.

Regarding the first group, FedCLAR and FedHome share the same aggregation mechanism as FedAvg, but they introduce unique elements to enhance the model's performance.
On the one hand, the models clustering adopted by FedCLAR allows to obtain a slight improvement over FedAVG in both HAR and Sleep Detection scenarios.
However, in HAR, clustering sometimes increases the loss value; an early sign of potential degradation, though without an immediate drop in accuracy.
Nevertheless, its final fine-tuning phase helps prevent a complete model overfitting, reducing loss and improving accuracy by about $20\%$ in HAR and 5–10\% in the other two scenarios.
Despite these gains, its overall performance remains significantly lower than that of FedSub and ProtoHAR.
On the other hand, the fine-tuning phase with synthetically balanced datasets used by FedHome yields noticeable gains over the other baselines in this group, reaching performance comparable to ProtoHAR in HAR.
Specifically, the FedHome personalization strategy enables up to 30\% improvement in HAR and approximately 10\% in the other two settings, leading to final performance similar to FedCLAR.

\begin{figure*}[t]
     \centering
     \begin{subfigure}[t]{0.32\textwidth}
         \centering
         \includegraphics[width=\textwidth]{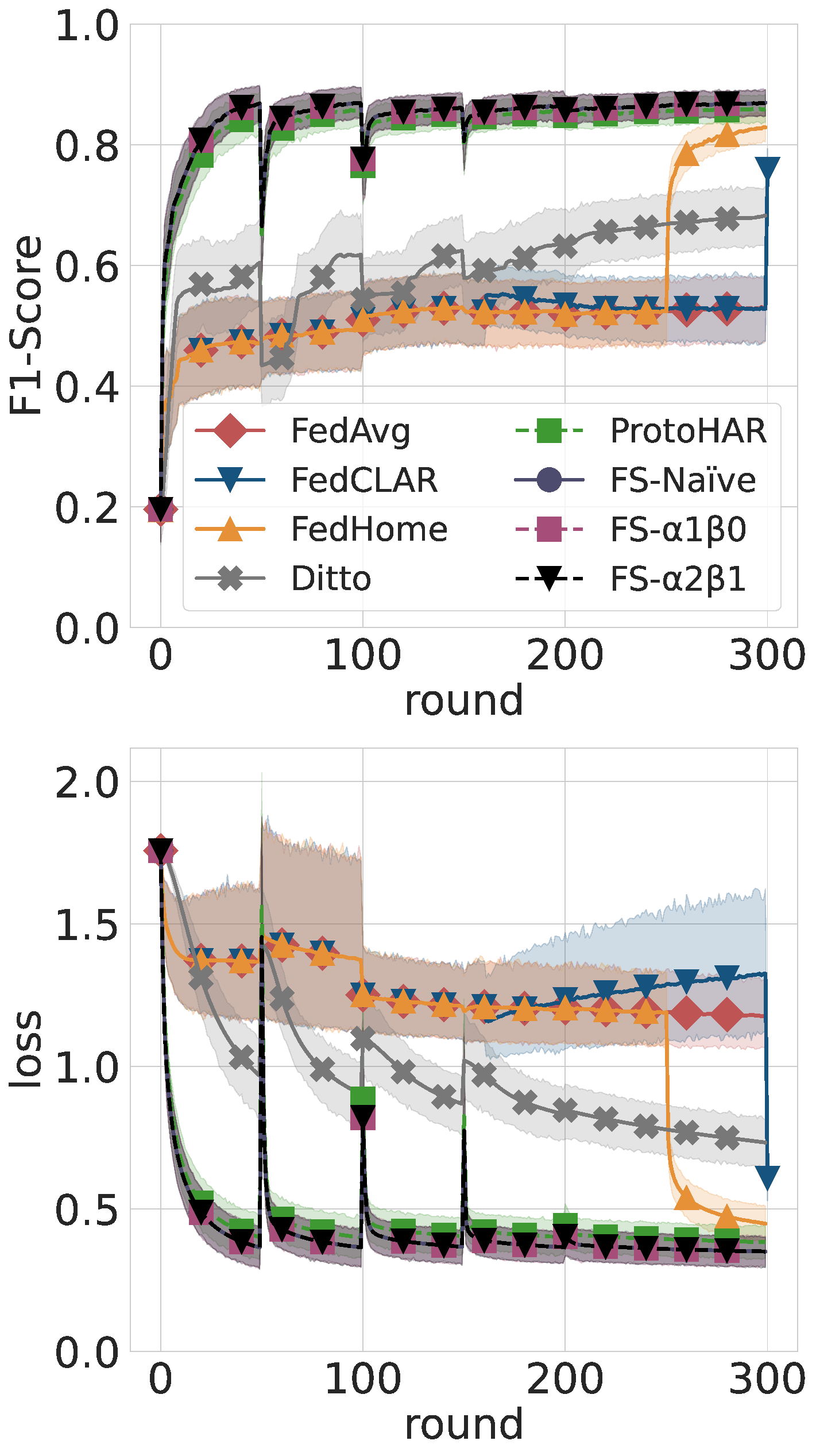}
         \caption{HAR}
         \label{fig:dynamic_wisdm}
     \end{subfigure}
     \begin{subfigure}[t]{0.32\textwidth}
         \centering
         \includegraphics[width=\textwidth]{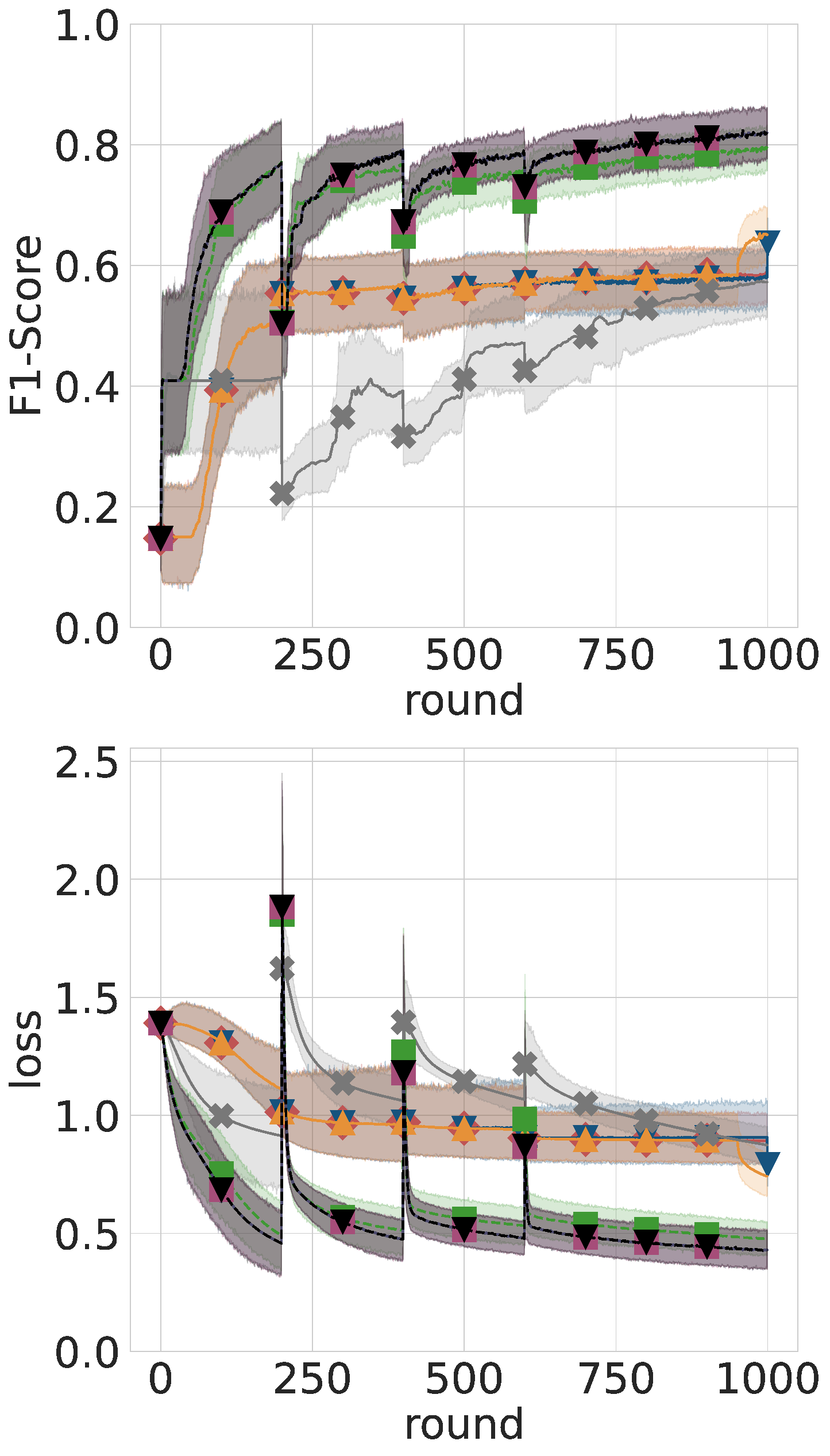}
         \caption{Stress Detection}
         \label{fig:dynamic_wesad}
     \end{subfigure}
     \begin{subfigure}[t]{0.32\textwidth}
         \centering
         \includegraphics[width=\textwidth]{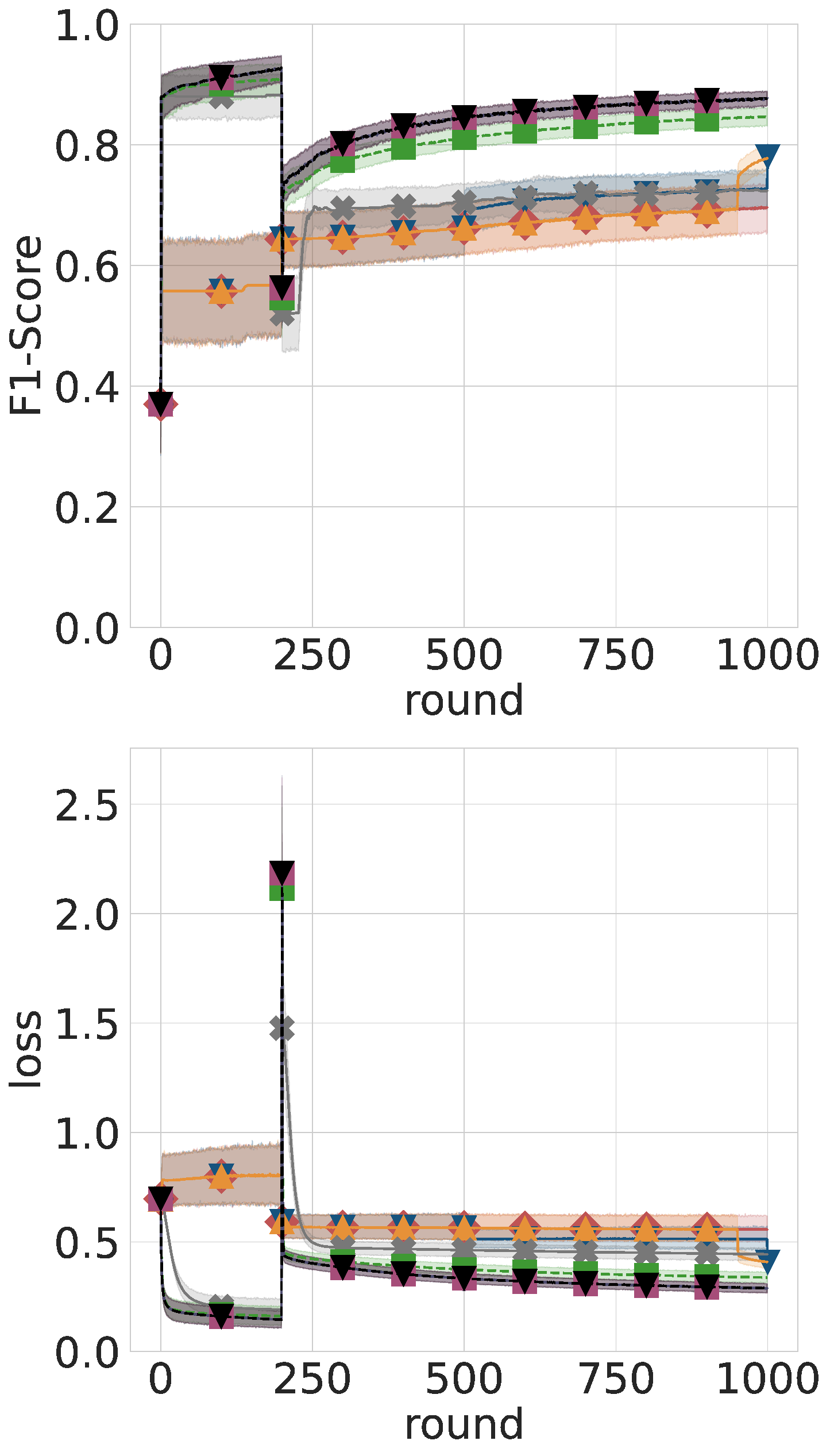}
         \caption{Sleep Detection}
         \label{fig:dynamic_dreamt}
     \end{subfigure}
    \caption{Average F1-Score and loss value with 95\% confidence intervals in the dynamic scenario.}
    \label{fig:dynamic_scenario}
\end{figure*}

Ditto shows limited improvements across all three scenarios, with performance consistently below FedCLAR and FedHome. Specifically, in HAR and Sleep Detection it performs relatively well in the early rounds but fails to match the final performance of FedCLAR and FedHome, converging to an F1-Score of $0.69$ and $0.73$, respectively.
In contrast, in Stress Detection it exhibits particularly slow convergence, eventually reaching a level comparable to FedAVG. This behavior can be explained by Ditto’s reliance on regularization toward the global model, which restricts the degree of personalization achievable at the client level. Such behavior is consistent with observations reported for other regularization-based approaches~\cite{9743558}, where strong data heterogeneity often causes suboptimal convergence and reduced accuracy.

Focusing on the highest-performing algorithms, both demonstrate strong performances in the initial rounds, quickly surpassing the other baseline. However, FedSub outperforms all the other approaches in every use case, reaching peak performance earlier than ProtoHAR and maintaining this advantage throughout all the communication rounds.
Specifically, in HAR, FedSub holds a $2\%$ F1-Score lead for the first $100$ rounds, then stabilizes at about $1\%$ until both reach maximum scores of $0.87$ and $0.86$.
Furthermore, in terms of test loss, FedSub maintains a $4\%$ advantage, reaching $0.34$ compared to ProtoHAR’s $0.38$.
In Stress Detection, the two algorithms start similarly; after round $200$, FedSub surpasses ProtoHAR, keeping a $2.83\%$ F1-Score and $6.23\%$ loss advantage, with final scores of $0.84$ in F1-Score and $0.39$ in loss value, while ProtoHAR obtains $0.81$ and $0.45$.
Finally, in Sleep Detection, FedSub's advantage over ProtoHAR is evident from the very first rounds, maintaining a 3.52\% lead in F1-Score and a 4.23\% lead in loss value for the first 600 rounds.
Although the gap slightly narrows in terms of F1-Score, FedSub outperforms ProtoHAR, achieving a final score of 0.89 and a loss value of 0.27, while ProtoHAR reaches 0.85 and 0.33, respectively.

Regarding the two strategies for subnetwork extraction implemented in FedSub, no significant differences are observed between the Na\"ive and LRP-based methods.
Specifically, in HAR, both methods yield identical results, but we can notice a slight difference in Stress Detection, where both LRP-based variants achieve a $1\%$ reduction in loss values, while in Sleep Detection, $LRP-\alpha2\beta1$ results in a $1\%$ improvement in both F1-Score and test loss.
The absence of major performance discrepancies between the two strategies underscores the effectiveness of the subnetwork fusion approach, which will be further analyzed in Section~\ref{sec:hyperparams} among the benefits of using the LRP-based solution to reduce the communication overhead between clients and the server.

\subsection{Dynamic scenario}

The dynamic scenario is designed to emulate the evolution and gradual generation of new data on the clients’ side during the FL process. Unlike static setups, where clients have full access to their entire dataset from the outset, this scenario introduces controlled data drift. Specifically, we randomly selected 60\% of the clients and removed up to 80\% of the classes from their local datasets before the start of the simulation. Then, at predefined intervals (every 50 communication rounds for HAR and every 200 rounds for Stress and Sleep Detection), each of these clients reintroduces one class, randomly chosen from those previously removed.
In this way, we aim to mimic the appearance of new patterns that were initially absent, such as a user in a HAR application beginning to perform a previously unrecorded activity like running or cycling.

The primary goal of this setup is to evaluate the ability of the considered FL algorithms to adapt effectively to the emergence of new data, studying how the models’ performance and stability react to such a dynamic event over time.
In particular, we expect algorithms like FedAvg, which average all client models, to be less affected by new data, as they inherently blend information from both existing and emerging patterns across clients.
On the other hand, personalized models like those generated by ProtoHAR and FedSub could potentially struggle with the sudden introduction of new data classes, leading to a more pronounced impact on their performance.

Figure~\ref{fig:dynamic_scenario} present the results obtained in this dynamic setup.
As anticipated, the first group of algorithms, including FedAvg, FedCLAR, and FedHome, is not significantly affected in the first two use cases; aside from minor variations, performance remains comparable to the static scenario.
However, in the Sleep Detection scenario, also this group of algorithms shows a noticeable performance drop of approximately 10\% in classification accuracy.
This decrease can be explained by the characteristics of the classification task, which differs from the other ones by including only two labels, making it particularly sensitive to changes in data distribution.
In fact, with only two classes, any shift in the data can have a larger relative impact on classification performance, as there is limited diversity in label representation for the model to adapt to new patterns.

Ditto exhibits marked instability in the dynamic scenario, with sharp drops in F1-Score whenever previously removed classes are reintroduced into the clients’ datasets. Indeed, unlike FedCLAR and FedHome that rely heavily on the global model, thus benefiting from knowledge of classes observed by other clients, the regularization approach implemented in Ditto leads to a local model that is more shaped by the client’s current distribution.
As a result, when a client encounters a reintroduced class, its model lacks prior exposure to it, and the partial information inherited from the global model is clearly insufficient to prevent a temporary performance collapse.
In contrast, ProtoHAR and FedSub achieve substantially higher overall classification performance, though they are particularly sensitive to new data introductions as Ditto.
However, contrary to the regularized approach, the two prototype-based solutions only experience short-term drops in their performance, which are rapidly recovered within just a few communication rounds.

It is important to emphasize that the performance of all algorithms does not collapse to zero during the observed drops, since classes are removed only for a random subset of clients, while the remaining clients continue to contribute updates for them throughout training.
This setting allows the global knowledge (whether in the form of regularized global guidance for Ditto, aggregated prototypes for ProtoHAR, or fused subnetworks for FedSub) to retain partial information about the missing classes. As a result, the performance degradation is mitigated and the recovery is accelerated once the data is reintroduced. Within this context, results indicate that although all methods eventually recover, ProtoHAR and FedSub maintain superior performance across all scenarios, with FedSub consistently adapting more rapidly than ProtoHAR and Ditto. In several cases, FedSub is able to regain or even surpass its pre-drift performance within only a few rounds, showing that its combination of robust classification accuracy and fast adaptation makes it particularly effective in managing dynamic environments.

\section{Sensitivity analysis}
\label{sec:hyperparams}

In this section, we analyze the impact of the most critical aspects of FedSub: subnetworks extraction and fusion strategies.
Understanding how these components influence both classification performance and communication overhead is crucial, especially in mobile-based applications where efficient computation and data transmission are key requirements.

Our previous experiments revealed that the Na\"ive approach and the two LRP-based variants achieve comparable classification performance across the three application scenarios.
However, these methods differ significantly in terms of the amount of information exchanged between clients and the server.
Specifically, while the Na\"ive approach includes in the extracted subnetwork all the model components activated during the forward pass, the goal of the LRP-based method is to retain only those parameters that are effectively relevant for the classification task, thus potentially optimizing the communication with the server.

\begin{figure}[t]
     \centering
     \includegraphics[width=0.9\columnwidth]{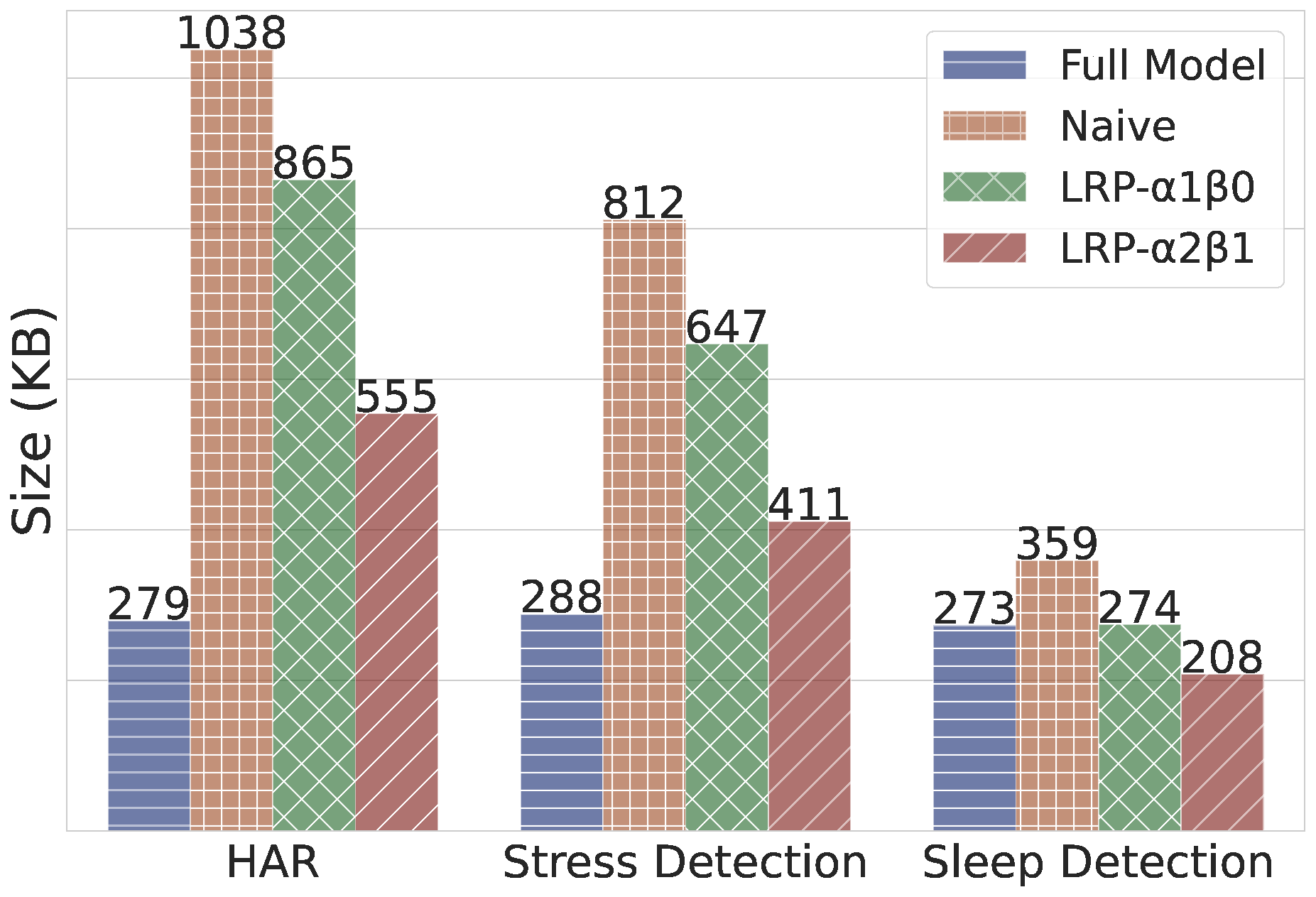}
     \caption{Average amount of data transmitted by clients in a FL round by using the three proposed subnetworks extraction strategies. \emph{Full Model} refers the standard FL approach of sending the whole local model to the server.}
     \label{fig:overhead}
\end{figure}

Figure~\ref{fig:overhead} presents the average amount of data sent by each client to the server per FL round across the different application scenarios. Here, \emph{Full Model} represents a standard FL approach in which the entire local model is transmitted, as seen in traditional methods such as FedAVG or FedHome.
In contrast, \emph{Na\"ive}, $\textit{LRP-}\alpha1\beta0$ and $\textit{LRP-}\alpha2\beta1$ indicate the average size of all the subnetworks transmitted by the clients while using FedSub and the proposed approaches to extract the class-level subnetworks.

From the results it is evident that \emph{Na\"ive} imposes the highest communication burden among the proposed methods. In the HAR scenario with six classes, it requires an average of 1.04 MB per client per round, which is 3.7 times the size of the full model transmission.
Similarly, in the Stress Detection scenario with four classes, the clients transmits, on average, 821 KB per round (2.8x the full model size), and in the Sleep Detection scenario with only two classes, it requires 359 KB (1.3x the full model size).
Although the amount of data exchanged via subnetworks does not grow linearly with the number of classes (suggesting an overlap between different subnetworks) the communication cost remains significantly higher than that of a traditional FL approach, which can limit the applicability of \emph{Na\"ive} in resource-constrained environments.

Among the LRP-based approaches, $\textit{LRP-}\alpha1\beta0$ demonstrates a notable reduction in communication overhead, allowing clients to send, on average, 173 KB less than \emph{Na\"ive} in HAR, 165 KB less in Stress Detection, and nearly matching the communication cost of a traditional FL solution in Sleep Detection.
However, the most efficient approach is $\textit{LRP-}\alpha2\beta1$, which further reduces the size of transmitted subnetworks by focusing only on the most essential components.
Specifically, this variant achieves an overhead of 1.9x in HAR, 1.4x in Stress Detection, and even transmits 65 KB less than \emph{Full Model} in Sleep Detection. As a result, $\textit{LRP-}\alpha2\beta1$ clearly emerges as a highly promising candidate for applications where upload bandwidth is a critical resource.

\begin{figure*}[t]
     \centering
     \begin{subfigure}[t]{0.29\textwidth}
         \centering
         \includegraphics[width=\textwidth]{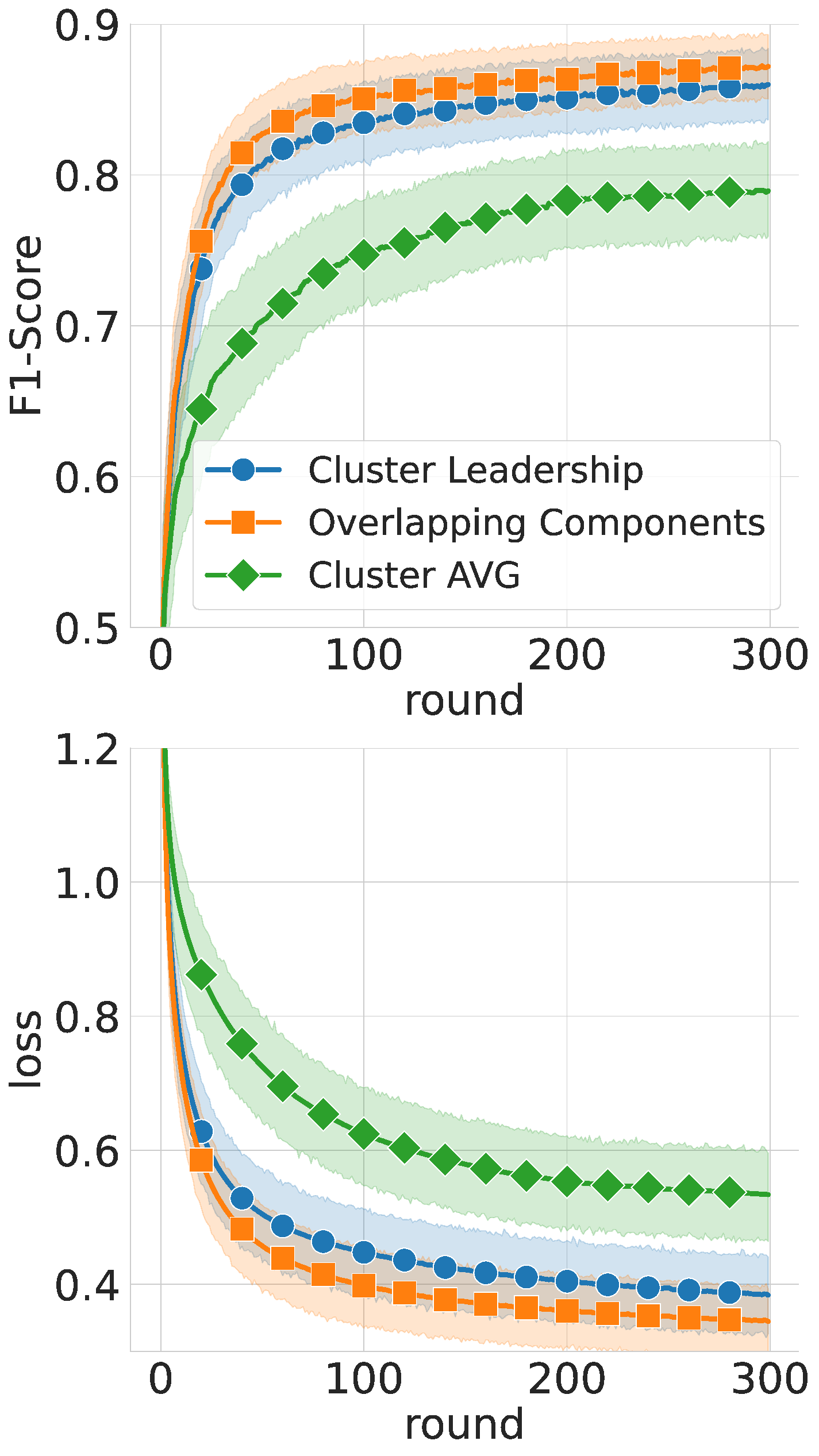}
         \caption{HAR}
         \label{fig:wisdm_merging_f1}
     \end{subfigure}
     \begin{subfigure}[t]{0.29\textwidth}
         \centering
         \includegraphics[width=\textwidth]{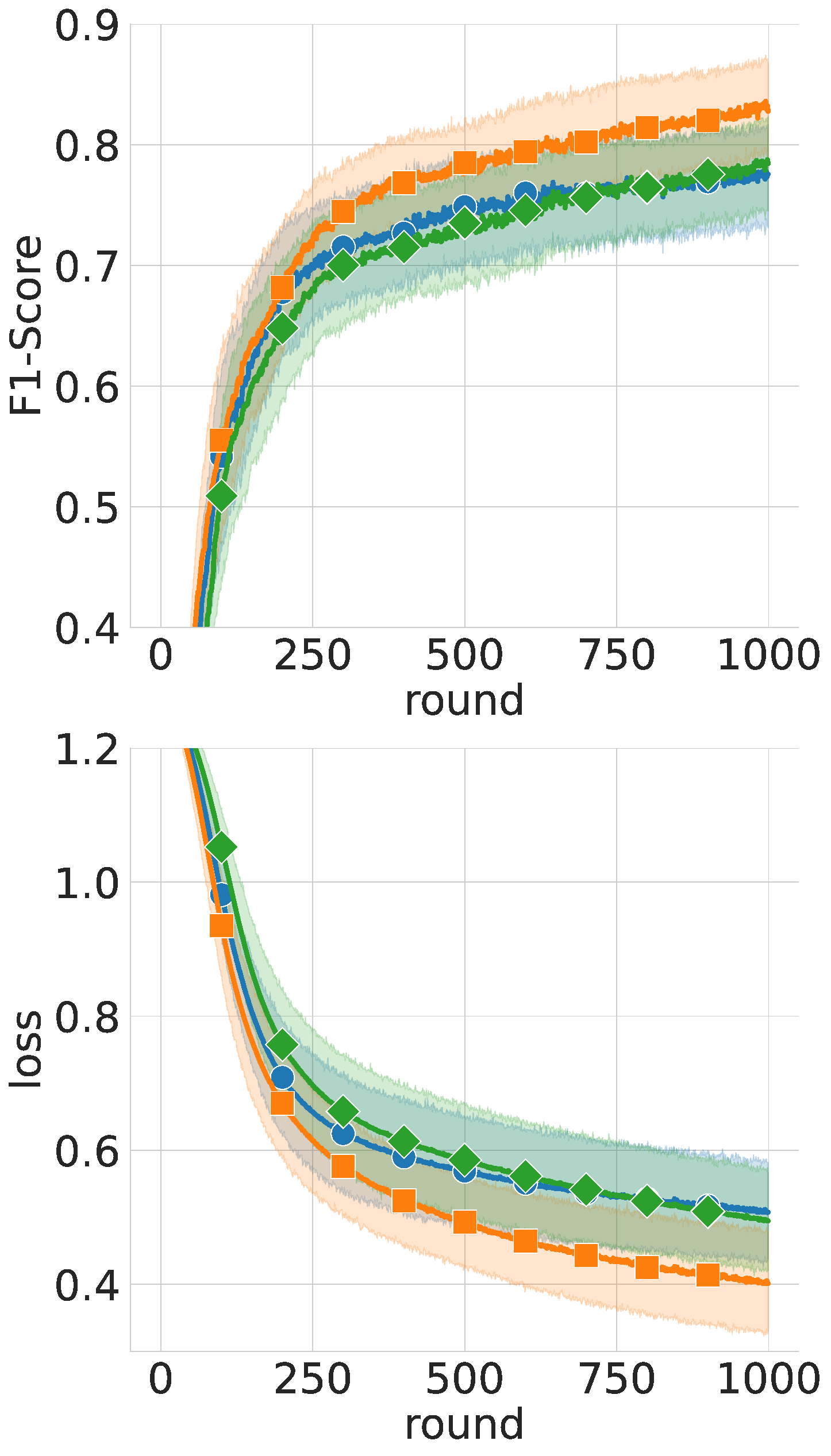}
         \caption{Stress Detection}
         \label{fig:wesad_merging_f1}
     \end{subfigure}
     \begin{subfigure}[t]{0.29\textwidth}
         \centering
         \includegraphics[width=\textwidth]{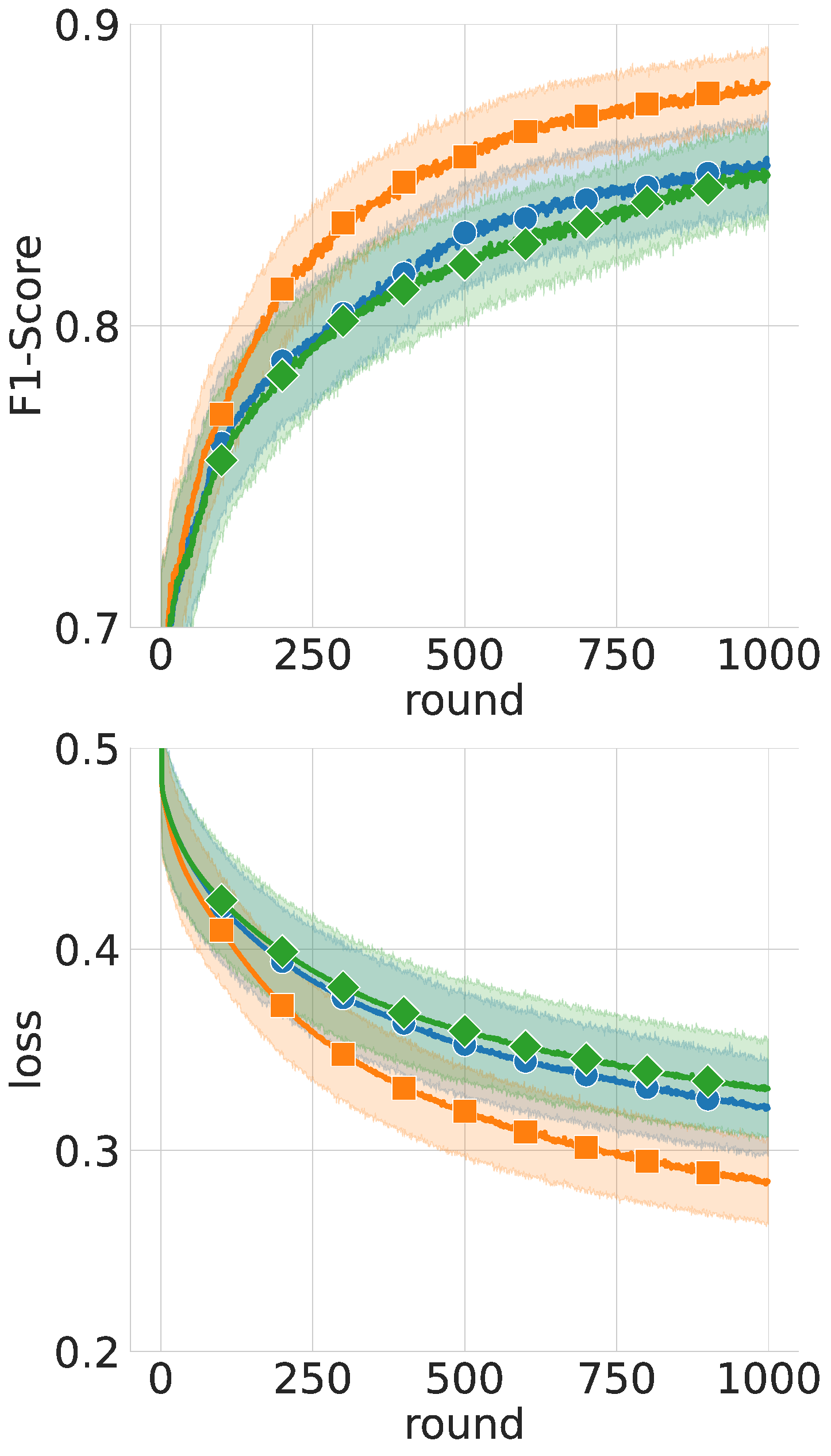}
         \caption{Sleep Detection}
         \label{fig:dreamt_merging_f1}
     \end{subfigure}
    \caption{Average F1-Score and loss value with 95\% confidence intervals for the different subnetworks fusion strategies.}
    \label{fig:merging_strategies_f1}
\end{figure*}

The second key aspect analyzed in this section is the impact of different fusion strategies on classification performance.
Figures~\ref{fig:merging_strategies_f1} shows the results in terms of F1-Score in the three application scenarios for the three proposed strategies.
The results clearly demonstrate that \emph{Overlapping Components} is the most effective fusion strategy in all scenarios, consistently achieving higher classification performance and faster convergence compared to the other two alternatives.

At the opposite end of the spectrum, \emph{Cluster AVG}, which is conceptually closer to the classic FL approach of FedAVG, performs the worst among the three strategies.
Specifically, it achieves values of $0.79$ in both HAR and Stress Detection, and $0.85$ in Sleep Detection, which correspond to performance degradations of $8\%$, $5\%$, and $3\%$, respectively, compared to \emph{Overlapping Components}, thus indicating that a simple averaging approach fails to leverage the personalized structure of the subnetworks effectively.
The third strategy, \emph{Cluster Leadership}, generally yields intermediate performance. In HAR, it trends towards the performance of \emph{Overlapping Components}, albeit with a slower convergence, ultimately reaching a score of $0.86$.
However, in Stress Detection and Sleep Detection, its behavior is more similar to \emph{Cluster AVG}, obtaining F1-Score of $0.78$ and $0.86$, respectively.
These findings highlight the superiority of \emph{Overlapping Components}, which ensures that subnetworks fusion is restricted to parameters that are common among users who exhibit similar behavioral patterns. This selective approach prevents unnecessary interference from less relevant model components, leading to faster convergence and improved overall performance. In contrast, strategies like \emph{Cluster AVG} and \emph{Cluster Leadership}, which apply broader and simpler aggregation mechanisms, struggle to maintain the balance between personalization and shared learning, ultimately resulting in lower classification performance.

\begin{figure}[t]
     \centering
     \includegraphics[width=0.97\columnwidth]{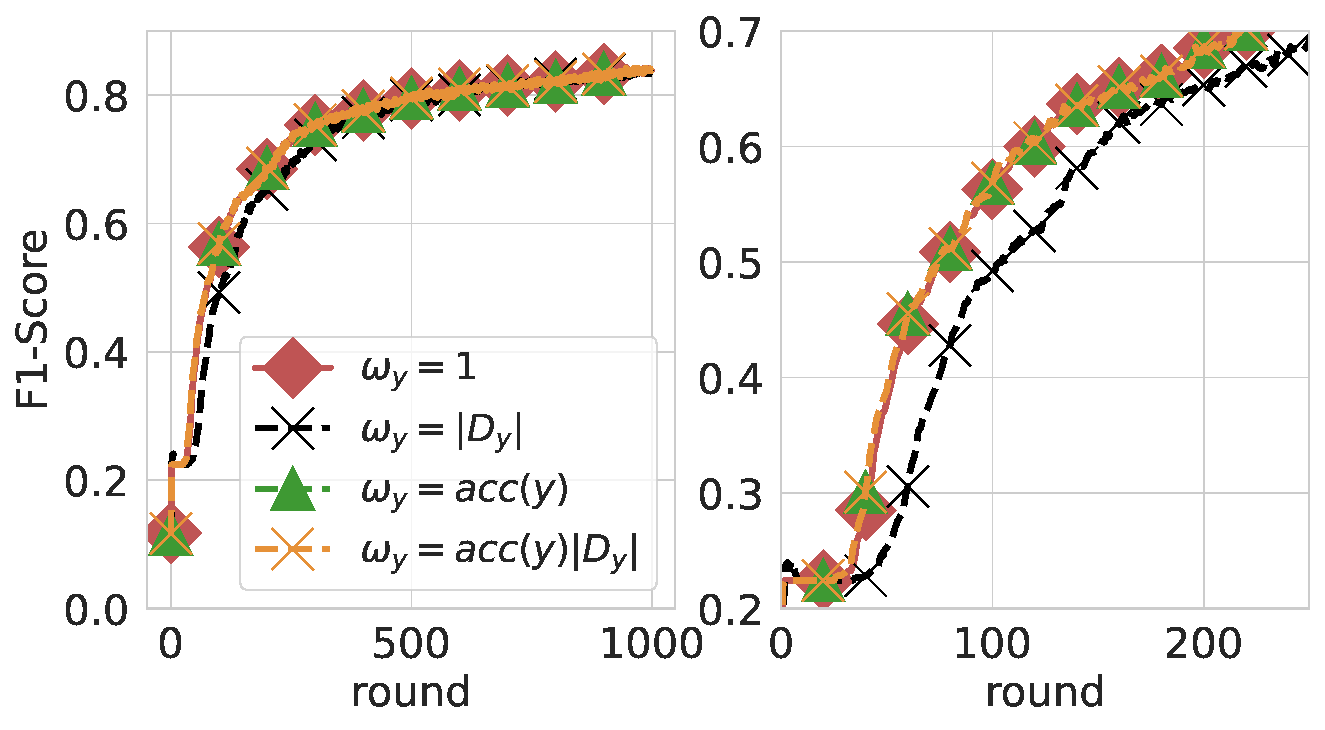}
     \caption{FedSub performance in the HAR scenario with different subnetwork relevance scores. The figure on the right highlights the differences in the first 250 communication rounds.}
     \label{fig:fedsub_weights}
\end{figure}

Finally, to further assess the impact of the subnetworks \emph{relevance score} on the fusion process, we evaluated FedSub with the Overlapping Components strategy under the different weighting criteria considered in our experiments.
The results obtained in the Stress Detection scenario (Figure~\ref{fig:fedsub_weights}) reveal that using $\omega_y = |D_y|$ may slightly slow down convergence during the early rounds, while the others allow for faster performance.
However, these differences are only transient: all configurations eventually converge to comparable levels of F1-Score, demonstrating that FedSub’s behavior is stable and not strongly dependent on the specific choice of relevance score.
This finding is further supported by similar trends observed in the Sleep Detection and HAR scenarios (not shown here for the lack of space), where the impact of different weighting strategies is minimal.
Overall, these results highlight the robustness and flexibility of FedSub, confirming its ability to deliver strong personalization regardless of the weighting mechanism adopted in the fusion phase.

\section{Conclusion and future work}
\label{sec:conclusions}

In this work, we introduced FedSub, a novel approach to Personalized Federated Learning that leverages class-aware prototypes clustering to guide fine-grained model aggregation through class-aware subnetwork fusion in order to address the challenges posed by heterogeneous data distributions.
Specifically, we proposed two different strategies to extract the most relevant subnetworks on the client side: (i) a \emph{Na\"ive} method, which includes all model components activated during the forward pass, and (ii) an approach based on Layer-wise Relevance Propagation (LRP-based), which identifies the most critical components by propagating the classification score back to the input and assigning a relevance score to each neuron in the model.

On the server side, these class-level subnetworks are then fused to generate personalized model updates for each client.
We designed and evaluated three different fusion techniques: (i) Cluster AVG, inspired by traditional FedAVG, which averages subnetwork parameters; (ii) Cluster Leadership, where the best-performing subnetwork guides model aggregation; and (iii) Overlapping Components, which averages only parameters that are common to all the participating subnetworks.

We provided a complexity analysis of FedSub, showing its ability to scale with the number of clients and its efficiency on the client, where the complexity of subnetwork extraction is comparable to standard forward and backward passes, making it suitable for resource-constrained devices.

Extensive experimental evaluations by simulating three real-world scenarios involving public datasets from mobile and wearable devices, allowed us to assess FedSub performance across diverse and challenging conditions characterized by high user data variability.
Our solution consistently outperformed state-of-the-art methods, achieving higher classification performance and faster convergence. Moreover, it proved its effectiveness in handling both static and dynamic data generation settings, even under significant data heterogeneity, demonstrating its potential for a wide range of applications.

Additionally, we conducted a sensitivity analysis to assess the impact of different subnetwork extraction and fusion techniques in terms of both classification performance and communication overhead. Although \emph{Na\"ive} and LRP-based extraction methods achieved comparable classification performance, the LRP-based techniques are particularly effective in reducing the amount of data transmitted from clients to the server. Regarding subnetwork fusion, the Overlapping Components strategy consistently outperformed the other two techniques across all three scenarios.

FedSub presents room for further optimization, particularly in the prototypes management. Indeed, in real-world scenarios, some clients may lack prototypes for certain classes if they have not yet generated data samples with those labels.
For instance, in a human activity recognition application, a user might not have performed a specific activity, or in a disease detection system, many users might be healthy and therefore lack examples for particular diseases.
To address this limitation, FedSub could be enhanced with a mechanism inspired by the Collaborative Filtering approach~\cite{Koren2022}, enabling the prediction of missing prototypes based on similarities among the data patterns of clients.
This extension could further improve FedSub's performance in dynamic settings by allowing the creation of class prototypes even before clients generate the corresponding data and share the associated prototype.
Anticipating prototype generation in this way would help to maintain robust model performance, especially in scenarios with highly heterogeneous and evolving data distributions.

Another avenue for improvement lies in optimizing the subnetworks extraction to further reduce communication overhead.
Although LRP significantly reduces the subnetwork's size compared to the \emph{Na\"ive} method, the amount of data transmitted by clients may still be problematic for applications where the uplink bandwidth represents a limited resource.
To address this issue, we plan to explore dedicated data compression techniques by grouping and transmitting shared parameters among different subnetworks only once, thus minimizing communication costs while preserving model performance.
Such optimization would allow us to deploy and test FedSub in scenarios with a higher number of labels, including real-world applications such as complex activity recognition, personalized recommender systems, and mobile health applications, where fine-grained distinctions between activities, preferences, or health states require a larger set of labels.
However, a major challenge in this direction is the scarcity of publicly available datasets that exhibit these characteristics, highlighting the need for new data collection initiatives to build comprehensive benchmarks to evaluate personalized federated learning approaches like FedSub.
In summary, our analysis shows that FedSub scales well in terms of computation, but its main limitation concerns communication efficiency in scenarios with very large label spaces, where the number of subnetworks may increase substantially.
In fact, while the proposed solution is already suitable for application scenarios such as HAR and mobile health, where the number of labels is typically limited, addressing this challenge will be crucial for deploying FedSub in settings involving hundreds or even thousands of fine-grained classes, thereby motivating our planned future work on communication-efficient subnetworks extraction strategies.

Another research direction is FedSub implementation in fully decentralized learning environments~\cite{10251949, PALMIERI2024110681}, in order to eliminate the reliance on a central server, which can represent a single point of failure and introduce privacy risks, as well as to distribute the computational burden across the clients.
While the current framework depends on a central server to coordinate model aggregation and client communication, transitioning to a peer-to-peer setting could improve scalability, enhance robustness, and strengthen privacy. In such a decentralized configuration, no central authority handles or aggregates user data, thereby reducing potential privacy vulnerabilities inherent to centralized systems.
Moreover, decentralized learning can enable faster model updates, as clients are no longer required to wait for synchronization with a central server. Instead, models or submodels can be exchanged and refined as soon as they are available, allowing for quicker adaptation to local data distributions.

Finally, such a decentralized paradigm can be further extended to opportunistic networks, where communication is established only when mobile devices come into proximity or fall within wireless range~\cite{9439130}. This setup could expand the applicability of FedSub to environments with intermittent or unreliable connectivity, including urban areas with frequent network disruptions and rural regions with limited communication infrastructure. By sharing only subnetworks instead of full models, FedSub has the potential to reduce communication overhead during the opportunistic contacts, making it particularly suitable for mobile and wearable devices operating under dynamic and resource-constrained conditions.

\bibliographystyle{IEEEtran}  
\bibliography{main.bib}

\end{document}